\documentclass[11pt]{article}

\usepackage{acl}

\usepackage{times}
\usepackage{latexsym}

\usepackage[T1]{fontenc}
\usepackage[utf8]{inputenc}

\usepackage{microtype}

\usepackage{inconsolata}

\usepackage{graphicx}

\usepackage{amsmath}
\usepackage{multirow}
\usepackage{rotating}
\usepackage{array}
\usepackage{tabularx}

\usepackage{hyperref}
\usepackage{url}
\usepackage{booktabs}

\usepackage{pdflscape}

\usepackage{amsmath,amsfonts,bm}

\def\eqref#1{equation~\ref{#1}}
\def\1{\bm{1}}

\def\vx{{\bm{x}}}

\def\vz{{\bm{z}}}

\def\mA{{\bm{A}}}

\def\mE{{\bm{E}}}

\DeclareMathAlphabet{\mathsfit}{\encodingdefault}{\sfdefault}{m}{sl}
\SetMathAlphabet{\mathsfit}{bold}{\encodingdefault}{\sfdefault}{bx}{n}

\definecolor{darkblue}{rgb}{0, 0, 0.5}
\hypersetup{colorlinks=true, citecolor=darkblue, linkcolor=darkblue, urlcolor=darkblue}
\newcolumntype{Y}{>{\raggedright\arraybackslash}X}
\newcolumntype{C}{>{\centering\arraybackslash}X}

\title{Through a PRISM, Clearly: Step-Sequence Reasoning for Mistake Recognition in Procedural Videos}

\hypersetup{
  pdftitle={Through a PRISM, Clearly: Step-Sequence Reasoning for Mistake Recognition in Procedural Videos},
  pdfauthor={Olga Loginova, Oleksandr Bezrukov, Frank Keller},
  pdfsubject={Procedural mistake recognition in videos},
  pdfkeywords={procedural video, mistake recognition, language grounding, semantic role labeling, hidden Markov model}
}

\author{
  Olga Loginova \\
  University of Trento \\
  Trento, Italy \\
  \texttt{olga.loginova@unitn.it}
  \And
  Oleksandr Bezrukov \\
  Gran Sasso Science Institute \\
  L'Aquila, Italy \\
  \texttt{oleksandr.bezrukov@gssi.it}
  \AND
  Frank Keller \\
  University of Edinburgh \\
  Edinburgh, United Kingdom \\
  \texttt{keller@inf.ed.ac.uk}
}

\begin{document}
\maketitle

\begin{abstract}
Procedural mistake recognition requires identifying deviations from valid multi-step activities. We study post-completion recognition from videos with benchmark-provided step boundaries. We introduce PRISM (Procedural mistake Recognition via Interpretable Semantic Modeling), which grounds video segments in language and maps their descriptions to procedural states using a lightweight hidden Markov model trained on correct executions. Reference alignment and semantic checks identify errors in step content and order. To support evaluation across heterogeneous benchmarks, we map dataset-specific annotations to four edit-based error types: insertion, deletion, substitution, and transposition. Across three benchmarks, we compare gold and VLM-generated descriptions in raw, semantic-role, and action–object forms. Decoder controls show consistent gains from the trained local decoder over nearest-candidate matching, with additional, dataset-dependent benefits from sequence decoding. Semantic role labeling improves Any-F1 in all six matched grounded comparisons, although the gains vary across datasets and do not translate into uniform improvements under separate configuration selection. These results establish language grounding as an effective interface for explicit procedural mistake reasoning and characterize the contributions of representation and decoding.
\end{abstract}

\section{Introduction}
\label{sec:intro}

A visually flawless plate of zoodles reaches the table, only for the first bite to reveal overboiled vegetables. The procedure appears finished before the failure becomes clear.

Some procedural errors become apparent only after task completion~\citep{Byrne2006Task, Reason2000HumanEM}. For AI systems monitoring or assisting human activity, the key question is retrospective: given the full video of an apparently completed task and a reference procedure, did the execution succeed, and if not, where did it go wrong? We focus on post-completion mistake recognition, which identifies errors from the complete recorded execution. This setting matters for troubleshooting, quality assurance, training, and human--AI assistance in real procedural environments~\citep{Norman2002TheDO, understandingmodels}.

Despite progress in temporal action segmentation and recognition \citep{ding2023temporalactionsegmentationanalysis, Ding2025TheJO, lai2024humanactionanticipationsurvey}, procedural mistake detection remains underexplored \citep{mazzamuto2025gazingmisstepsleveragingeyegaze, bacharidis2025visionbasedmistakeanalysisprocedural}. Existing approaches are fragmented: they are often vision-first and use different error labels, annotation units, and evaluation protocols \citep{peddi2024captaincook4ddatasetunderstandingerrors, haneji2025egooopsdatasetmistakeaction, Lee_2025_ICCV}. Recent work on procedural error taxonomies also highlights the need for shared categories across datasets~\citep{Loginova_2026_CVPR}. These differences complicate comparisons across datasets and obscure the contributions of grounding, representation, and procedural reasoning.

\begin{figure}[t]
\begin{center}
\includegraphics[width=\columnwidth]{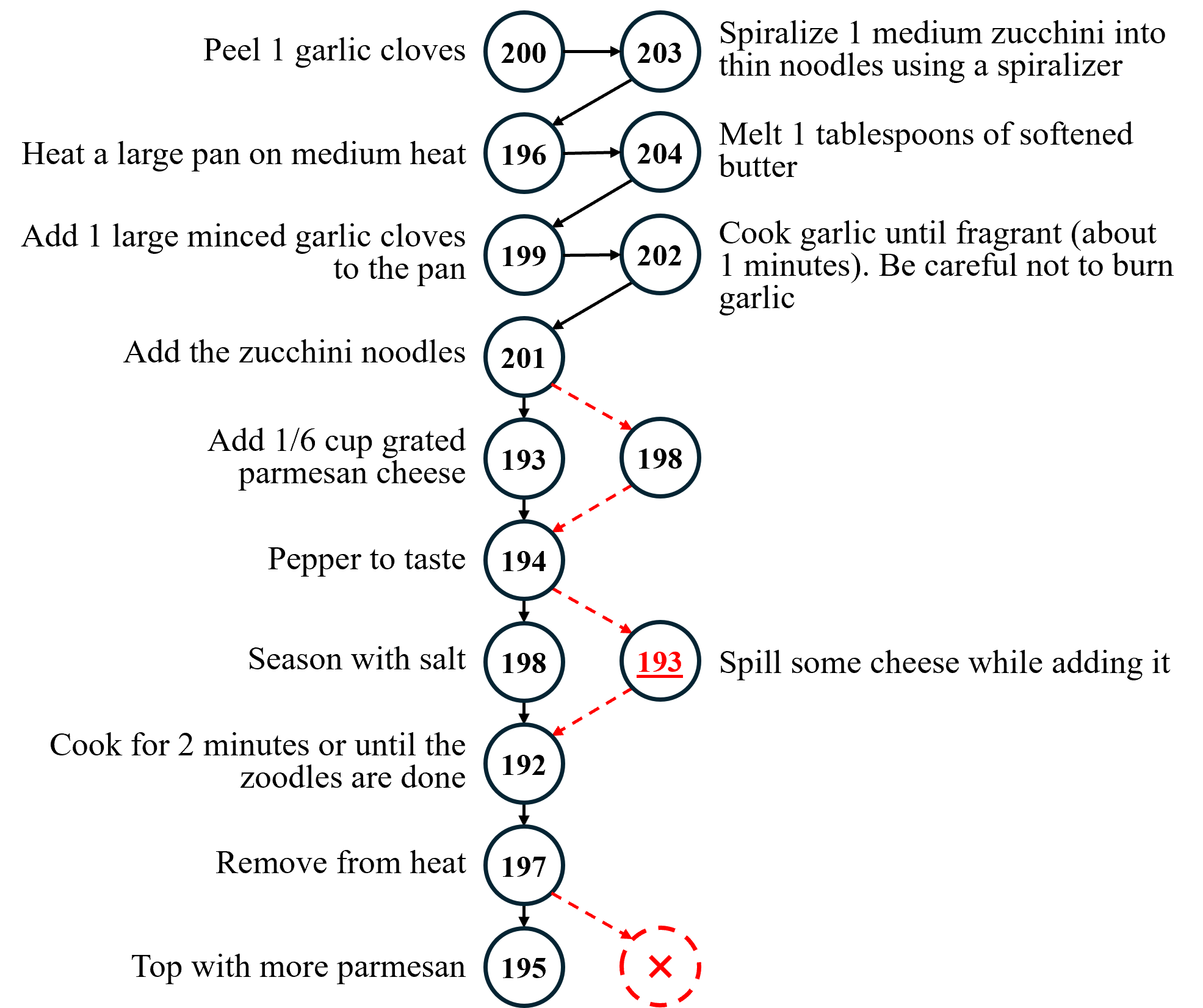}
   \end{center}
   \caption{PRISM representation of the valid Zoodles procedure and an erroneous execution from CaptainCook4D. Circles contain Step IDs, adjacent text gives step descriptions, and solid arrows trace the reference order. Red marks show a transposition of Steps 193, 194, and 198, a substitution at Step 193, and the deletion of Step 195.}
   \label{fig:tasks}
\end{figure} 

Following the procedural video taxonomy of \citet{bacharidis2025visionbasedmistakeanalysisprocedural}, we distinguish temporal localization from logical error identification after action completion. Temporal segmentation and action recognition determine what happened and when. We study the subsequent reasoning problem: given step-level evidence and a reference procedure, does the completed execution deviate from the protocol, and where? 

We evaluate this reasoning problem on EgoPER~\citep{Lee_2024_CVPR}, CaptainCook4D~\citep{peddi2024captaincook4ddatasetunderstandingerrors}, and EgoOops~\citep{haneji2025egooopsdatasetmistakeaction}. Related systems include EgoPED~\citep{Lee_2024_CVPR}, AMNAR~\citep{huang2025modelingmultiplenormalaction}, and AEM~\citep{guo2026proceduralmistakedetectionaction}. Their temporal input and evaluation protocols are discussed in Section~\ref{sec:exp}. We model this reasoning problem over discrete step semantics: procedural errors depend on what step was performed, with which arguments, and in what order.

This view motivates our language-grounded framework, because these constructs are naturally expressed in language. We propose a \emph{video $\rightarrow$ language $\rightarrow$ reasoning} pipeline. 

Like an optical prism refracting raw light into its constituent colors, our PRISM (Procedural mistake Recognition via Interpretable Semantic Modeling) makes procedural structure explicit through steps, semantic roles, and their order.\footnote{Code will be released upon acceptance.} Benchmark-provided segment boundaries isolate mistake recognition from temporal segmentation. For each segment, a vision-language model (VLM) receives the clip and the task's candidate descriptions as context, then generates a free-form step description. PRISM encodes this description as raw text or as a semantic role representation produced through semantic role labeling (SRL). A hidden Markov model (HMM) maps the representation to a task-local step identifier (Step ID). Edit alignment identifies deviations from a valid reference sequence. We reconcile the benchmark-specific annotation schemes through a common taxonomy of four edit operations: insertion, deletion, substitution, and transposition. Figure~\ref{fig:tasks} contrasts a valid procedure with an erroneous execution under this taxonomy. Circles denote HMM states labeled by Step IDs. The adjacent step descriptions are encoded as observations, and the ordered state sequence represents the procedure.

% We address three research questions: (1) Which textual representation best supports mistake recognition? (2) Does our unified framework generalize across datasets? (3) How reliable are modern vision-language models (VLMs) for video-to-text grounding under a common post-completion protocol?

% We compare benchmark step text, SRLs, and action--object abstractions, and study both the open model Qwen 3.5-27B~\citep{qwen3.5} and the closed model Gemini 3 Flash~\citep{google2025gemini3} as video-to-text grounding modules. We evaluate our approach on the datasets CaptainCook4D \citep{peddi2024captaincook4ddatasetunderstandingerrors}, EgoOops \citep{haneji2025egooopsdatasetmistakeaction}, and EgoPER \citep{Lee_2024_CVPR}. Our results show that language-grounded reasoning is a strong alternative to vision-centric methods, and that semantic normalization with SRLs significantly improves robustness against noisy VLM predictions.
%

Our contributions are:
\begin{itemize}
\item \textbf{PRISM, a language-grounded framework} that combines a lightweight HMM trained on correct executions with reference alignment and semantic checks to identify procedural mistakes.

\item \textbf{Cross-benchmark error analysis} through a shared edit representation of procedural deviations.

\item \textbf{A controlled study of step representations and decoding}, comparing gold and VLM-generated descriptions, matched Raw–SRL configurations, action–object abstractions, and independent versus transition-aware inference.
\end{itemize}

%
% We show that procedural mistakes can be detected effectively after video has been translated into language. This perspective turns varying, benchmark-specific error analysis into a unified reasoning problem over step sequences and provides a compact, interpretable path toward post-completion procedural mistake detection in video.
\section{Related work}
\label{sec:rel}

\paragraph{Mistake-aware procedural video benchmarks.}
Despite progress in procedural video understanding, only a small subset of benchmarks explicitly annotates mistakes.
Current benchmarks span cooking \citep{peddi2024captaincook4ddatasetunderstandingerrors, Lee_2024_CVPR}, assembly \citep{Sener2022Assembly101ALA, schoonbeek2023industrealdatasetprocedurestep, Jang2019EPICTentAE, Wang2023HoloAssistAEA}, laboratory tasks \citep{qian2022svipsequenceverificationprocedures}, and mixed scenarios \citep{haneji2025egooopsdatasetmistakeaction} but suffer from heterogeneous error definitions, annotation levels (video vs. step), and description styles (free-form vs. labels). \citet{Loginova_2026_CVPR} similarly argue that procedural video datasets need shared categories for mistakes and corrections across tasks and domains. Our experiments span CaptainCook4D, EgoPER, and EgoOops: CaptainCook4D contains 384 videos across 24 cooking tasks; EgoPER concentrates 386 videos in five cooking tasks; and EgoOops adds 50 videos from five mixed-domain procedures. This combination tests VLM grounding and PRISM under broad task coverage, dense repetition within tasks, and limited data outside cooking.

\paragraph{Mistake recognition in procedural videos.}
Prior approaches are predominantly vision-centric, modeling deviations from video features via recognition, anticipation, action-effect modeling, or structured prediction \citep{Lee_2024_CVPR, peddi2024captaincook4ddatasetunderstandingerrors, Lee_2025_ICCV, mazzamuto2025gazingmisstepsleveragingeyegaze, huang2025modelingmultiplenormalaction,guo2026proceduralmistakedetectionaction}. Recent work also explores interpretability in error reasoning \citep{storks2025transparentcoherentproceduralmistake}. Some work focuses on online detection and intervention forecasting using modified datasets \citep{Flaborea2024PREGOOMA, Plini2024TIPREGOCOA, Wang2023HoloAssistAEA, seminara2025differentiabletaskgraphlearning}. We target post-completion recognition, similarly to \citet{huang2025modelingmultiplenormalaction, narasimhan2023learningverificationtaskstructure, 9897369, Ding2023EveryMCA, haneji2025egooopsdatasetmistakeaction, li2025mistakeattributionfinegrainedmistake}. We perform core reasoning in the language space by grounding video executions into textual step sequences.

\paragraph{Semantic roles and structured step representations.}
Semantic role labeling provides a structured interface for events through predicate--argument structures \citep{baker-etal-1998-berkeley-framenet, gildea-jurafsky-2002-automatic, palmer-etal-2005-proposition, chen2025semanticrolelabelingsystematical}. In multimodal settings, semantic structure supports video understanding \citep{sadhu2021visualsemanticrolelabeling}, procedural instruction tracking \citep{batra-etal-2025-predicting}, and fine-grained mistake attribution \citep{li2025mistakeattributionfinegrainedmistake}. 

We use SRLs as an intermediate normalization layer to bridge free-form step text and mistake reasoning.

\begin{figure*}[t]
  \centering
   \includegraphics[width=1\linewidth]{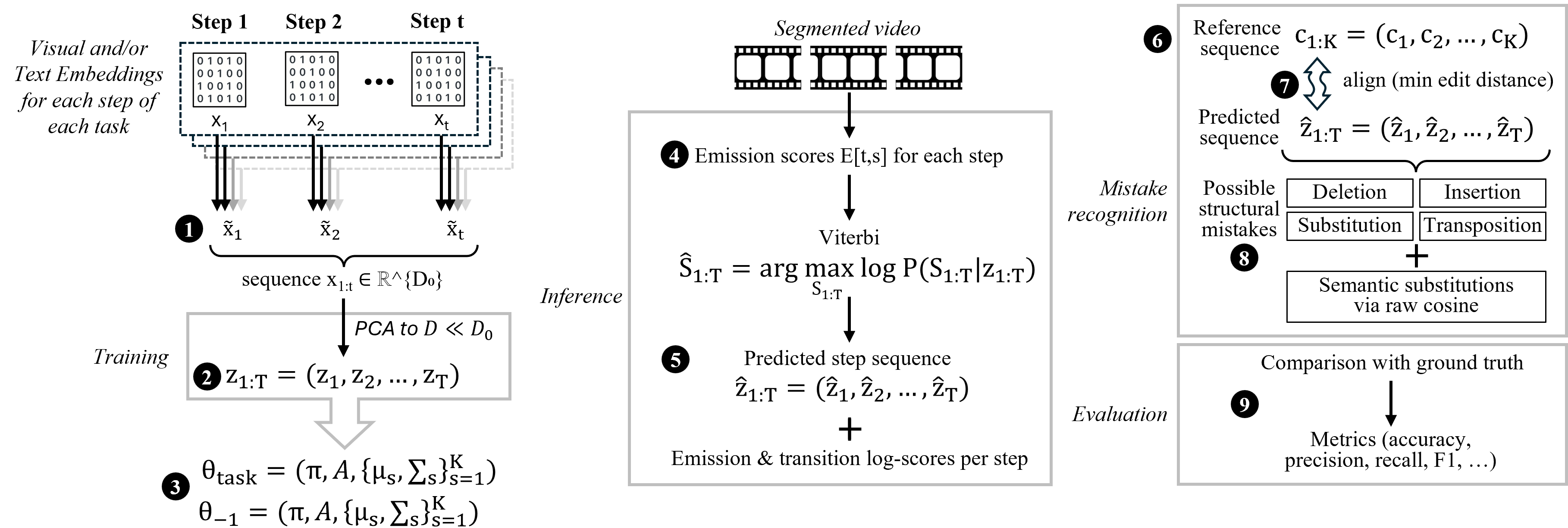}

   \caption{The PRISM pipeline for post-completion mistake recognition. (1)~For each step of each training video, we obtain a textual, visual, or multimodal encoding,~$\vx_t$. (2)~After PCA regularization, we obtain the representation of each step,~$\vz_t$. (3)~During training, the model $\theta_{\text{task}}$ learns from sequences of $\vz_t$ for each task. Additionally, when we put all tasks together, we obtain a global HMM $\theta_{-1}$. (4)~At inference time, we obtain embeddings for each video segment in the same way as during training and compute the emissions $[\bm{E}]_{t,s}$. (5)~We then run Viterbi for sequence inference and keep the emission and transition scores for each step. (6)~For each task, we also have a set of possible non-erroneous sequences. (7)~We align the predicted sequence to the reference sequence using the selected sequence-alignment method. (8)~Mismatches in the alignment indicate possible structural errors. Falling below the predefined thresholds for emission and transition scores indicates possible low-probability events. (9)~During evaluation, we compare the predicted errors with the gold-standard error annotations of the dataset.}
   \label{fig:hmm}
\end{figure*}
\section{Methodology}
\label{sec:met}

Figure~\ref{fig:hmm} summarizes PRISM. Intuitively, our HMM captures two complementary aspects of a correct procedure: what a valid step looks like (emission) and how steps logically progress (transition), learning both exclusively from successful executions. 

Each task provides candidate step descriptions, valid reference sequences, and correct training executions. Candidate steps define the task-local state inventory. We encode each segment using the representations in Section~\ref{sec:vid_to_txt} and estimate HMM parameters from correct executions. At inference, the HMM decodes the executed step sequence. Reference alignment identifies structural edits, while emission, transition, and semantic similarity scores identify atypical observations and transitions. Appendix~\ref{app:hmm} reports the configuration-specific settings.

The grounding interface, SRL schema, text-encoding stage, HMM estimator and decoder, and edit-alignment procedure remain unchanged across tasks. Each procedure supplies its task-local candidate inventory and reference sequences, while its HMM parameters are estimated from correct executions.

\subsection{HMM training}
\label{sec:hmm_tr}

Each video is represented as a sequence of pre-segmented step clips. Section~\ref{sec:vid_to_txt} defines the textual, SRL, action--object, visual, and multimodal representations used for each clip. For a video with \(T\) clips, let
\[
\bm{x}_{1:T}=(\bm{x}_1,\dots,\bm{x}_T),
\qquad
\bm{x}_t\in\mathbb{R}^{D_0}.
\]
When specified by the configuration, we L2-normalize the embeddings and add Gaussian jitter to a random subset of training embeddings. We fit principal component analysis (PCA) on the training embeddings and obtain
\[
\bm{z}_t=\operatorname{PCA}(\bm{x}_t)\in\mathbb{R}^{D}.
\]

For task \(a\), the hidden-state set
\(\mathcal{S}_a=\{1,\dots,K_a\}\) contains its Step IDs. The projected embeddings \(\bm{z}_{1:T}\) are the observations. The model parameters are
\[
\theta_a=
\left(
\boldsymbol{\pi},
\bm{A},
\{\phi_s\}_{s\in\mathcal{S}_a}
\right),
\]
where \(\boldsymbol{\pi}\) is the initial-state distribution, \(\bm{A}\) is the transition matrix, and \(\phi_s\) contains the emission parameters for state \(s\). For Gaussian emissions, \(\phi_s=(\boldsymbol{\mu}_s,\boldsymbol{\Sigma}_s)\). We estimate \(\boldsymbol{\mu}_s\) from the state observations and \(\boldsymbol{\Sigma}_s\) with Ledoit--Wolf shrinkage. For cosine emissions, \(\boldsymbol{\mu}_s\) serves as the state prototype.

For the transition model, let \(c_{ss'}\) denote the number of times the transition \(s \to s'\) is observed in the training set. We then apply Laplace smoothing with parameter \(\beta\):
\begin{equation}
A_{ss'} = \frac{c_{ss'} + \beta}{\sum_{k=1}^{K_a} (c_{sk} + \beta)}.
  \label{eq:7}
\end{equation}
After Laplace smoothing, we apply local transition constraints based on the valid reference sequences by suppressing out-of-band transitions and renormalizing each row. We estimate the initial state distribution in the same way. Let \(m_s\) be the number of correct training videos whose first step is \(s\). We then define $\pi_s = \frac{m_s + \beta}{\sum_{j=1}^{K_a} (m_j + \beta)}$. Thus, both the start distribution and the transition matrix are smoothed empirical estimates from correct executions.

% In addition to per-task HMMs, we train a global model $\theta_{-1}$ by pooling data across all tasks. This model is used as a fallback when task-specific parameters are unavailable.

During training we record calibration statistics (mean, quantiles) of emission log-likelihoods and transition log-probabilities on correct videos.
Let \(\bm{\delta}_{t,s}=\bm{z}_t-\boldsymbol{\mu}_s\). The log emission likelihood is
\begin{equation}
\begin{aligned}
\log p(\bm{z}_t \mid S_t=s)
&= -\frac{1}{2}\Bigl[
\bm{\delta}_{t,s}^\top \boldsymbol{\Sigma}_s^{-1}\bm{\delta}_{t,s} \\
&\quad + \log |\boldsymbol{\Sigma}_s| + D\log(2\pi)
\Bigr].
\end{aligned}
\label{eq:25}
\end{equation}
%
% Here the term 
% \((\bm{z}_t - \boldsymbol{\mu}_s)^\top \boldsymbol{\Sigma}_s^{-1} (\bm{z}_t - \boldsymbol{\mu}_s)\)
% is the squared Mahalanobis distance between \(\bm{z}_t\) and the state mean.
These standard log-likelihood statistics are stored for analysis and threshold tuning.

The term \(\log \lvert \boldsymbol{\Sigma}_s \rvert\) penalizes the volume of the Gaussian, and \(D \log (2\pi)\) is the normalization constant ensuring that the Gaussian density integrates to 1.

\subsection{HMM inference}
\label{sec:hmm_inf}

At test time, the model extracts, normalizes, and projects step-level features $\bm{x}_t$ into $\bm{z}_t$ identically to training.

We then compute an emission score matrix $\mE \in \mathbb{R}^{T \times K}$, where $E_{t,k}$ is the emission score assigned to observation $\bm{z}_t$ under state $k$. For Gaussian emissions, we compute $E_{t,k} = \log \mathcal{N}(\bm{z}_t; \boldsymbol{\mu}_k, \boldsymbol{\Sigma}_k)$. 

We also evaluate cosine-based emissions for normalized embeddings. In that case,
$E_{t,k} = T_{\cos}\,\cos(\bm{z}_t, \boldsymbol{\mu}_k)$, where \(T_{\cos} > 0\) is a temperature parameter.

This cosine form matches the geometry of embedding spaces trained or commonly used with angular similarity, and can be interpreted as an unnormalized state score. We L2-normalize vectors before cosine to obtain a bounded, temperature-scaled score.

We also support a global emission scaling factor $\alpha$, so that $[\tilde{\bm{E}}]_{t,k} = \alpha [\bm{E}]_{t,k}$. This adjusts the relative weight between emission and transition terms in the Viterbi objective. 
% In the Gaussian case, $E_{t,k}$ is a log-likelihood; in the cosine case, it is a temperature-scaled similarity score.

% For cosine emissions we compute $E_{t,k} = T \cdot \cos(z_t, \mu_k)$, with temperature $T>0$. This cosine form matches the geometry of normalized CLIP-like features and can be seen as a logit for a softmax over states. We L2-normalize vectors before cosine to ensure a bounded, temperature-scaled score.

We then use the Viterbi
% Viterbi is a dynamic-programming algorithm that finds the highest-scoring hidden-state sequence under the chosen start, transition, and emission scores.}
algorithm~\citep{1054010} to decode the highest-scoring step sequence $\hat{S}_{1:T} = \arg\max_{S_{1:T}} \mathrm{Score}(S_{1:T}, \bm{z}_{1:T})$. The path score is
\begin{equation}
\begin{aligned}
\mathrm{Score}(S_{1:T}, \bm{z}_{1:T})
&= \log \pi_{S_1}
  + \sum_{t=2}^{T} \log A_{S_{t-1},S_t} \\
  &\quad
  + \sum_{t=1}^{T} \tilde{E}_{t,S_t}.
\end{aligned}
\label{eq:23}
\end{equation}
For Gaussian emissions, this reduces to a sum of start, transition, and emission log-likelihood terms.

After predicting the sequence, we record for each time step $t$ the emission score
$\ell_t^{\mathrm{em}} = \tilde{E}_{t,\hat{S}_t}$, and for each step $t>1$ the transition score
$\ell_t^{\mathrm{tr}} = \log A_{\hat{S}_{t-1}, \hat{S}_t}$. We refer to these as the HMM scores along the Viterbi path.

\subsection{Step alignment and mistake recognition}
\label{sec:pcmd_module}

Each procedure, regardless of its specific execution in the dataset, has one or more valid correct step sequences. This is an abstract representation of the different ways to successfully complete the task. The pool of such reference sequences is formed either from the corresponding written instructions provided by annotation guidelines or from error-free videos. Let a reference sequence for a given procedure be $\bm{r}_{1:L} = (r_1,\dots,r_L)$, where each \(r_\ell\) is a step ID and \(L\) is the length of that reference sequence.

Next we compare the sequence predicted by the algorithm $\hat{S}_{1:T}$ with one of the reference sequences, namely, the closest reference variant under the selected alignment method. We align the predicted sequence to the chosen reference sequence using global sequence alignment.\footnote{We use either Needleman--Wunsch or difflib. Needleman--Wunsch is a classical dynamic-programming algorithm for global sequence alignment. It optimizes a well-defined objective and is preferable when we need full control over substitution and gap penalties. Python's difflib.SequenceMatcher is a heuristic matcher that focuses on maximizing the length of matching blocks. 
%It is often more forgiving and may produce alignments that match human intuition when sequences share long common subsequences but differ locally. It is fast and works well as a robust default when we do not want to carefully tune alignment costs. In practice, we use difflib as the default alignment mode.
}
This alignment identifies structural errors: \textbf{insertions} (extra steps), \textbf{deletions} (missing steps), \textbf{transpositions} (wrong order of correct steps)\footnote{We identify a transposition by pairing a deletion and an insertion of the same step ID at different alignment positions.}, and \textbf{substitutions} (mismatched steps at an aligned position). 

In addition to structural errors, we use the Viterbi-path scores directly. Low values of \(\ell_t^{\mathrm{em}}\) indicate atypical observations under the assigned state, and low values of \(\ell_t^{\mathrm{tr}}\) indicate atypical state transitions. 
% Their negated versions $s_t^{\mathrm{em}} = -\ell_t^{\mathrm{em}}$ and
% $s_t^{\mathrm{tr}} = -\ell_t^{\mathrm{tr}}$ serve as anomaly scores at step \(t\) for emission and transition respectively. 
Given the configuration-specific thresholds
\(\tau_{\mathrm{em}}\) and \(\tau_{\mathrm{tr}}\), reported in Appendix~\ref{app:hmm}, we flag a LowProbEmission if
\(\ell_t^{\mathrm{em}} < \tau_{\mathrm{em}}\), 
and for \(t>1\) a LowProbTransition if
\(\ell_t^{\mathrm{tr}} < \tau_{\mathrm{tr}}\). These two classes provide score-based diagnostics that complement the edit labels:
LowProbEmission flags an observation that is unlikely under its decoded step state,
while LowProbTransition flags an unlikely adjacent step transition in the decoded path. They can overlap with edit-based errors, but they also identify low-confidence or locally implausible events that are not cleanly summarized by a single insertion, deletion,
substitution, or transposition\footnote{LowProbEmission flags a step whose decoded ID is correct but whose visual or textual content is atypical, such as using the wrong quantity, tool, or destination. For example, in EgoPER, the grounded text \textit{Drop tortilla on floor}
is decoded as the procedural state \textit{Place tortilla on cutting board}, but receives a very low emission score. LowProbTransition can flag an unlikely local ordering even when both neighboring steps are individually valid, e.g., a CaptainCook4D transition
from rinsing mushrooms directly to adding them to the pan before the expected
drying, stemming, and slicing steps.}.

Structural alignment cannot expose an altered execution when the predicted and reference steps share the same Step ID. In this case, we compute the cosine similarity between the observation and the mean embedding of its assigned state. A similarity below \(\tau_{\mathrm{sem}}=0.85\) produces a semantic Substitution. This rule captures changes in quantities, tools, destinations, and other arguments hidden by a shared Step ID.

\subsection{Language grounding and step representations}
\label{sec:vid_to_txt}

We use language as the interface between video and procedural reasoning. We call benchmark-provided or benchmark-derived step descriptions gold text and VLM-generated descriptions grounded text. Gold-text runs bypass video grounding and do not use video. Grounded runs use descriptions generated by a VLM. The HMM decodes the resulting embedding sequence into task-local Step IDs.

\paragraph{Gold step text.} We start from the canonical step descriptions provided by each benchmark. For erroneous steps, we use the benchmark-specific observed or modified step descriptions when they are available. When a benchmark provides only a short error cue instead of a full observed step, we convert it into a complete modified description by combining the canonical step text with the error cue. We also apply minimal text cleanup to remove benchmark-specific artifacts that do not carry procedural meaning. Appendix~\ref{app:dataset_profiles} gives representative dataset-specific examples.

The resulting representation keeps the lexical detail of the original annotation, including quantities, destinations, surfaces, and local modifiers. That detail matters because many procedural mistakes differ from the canonical step in terms of the arguments, not only in terms of the main verb.

\paragraph{Semantic role representations.}
We convert each step description into a predicate--argument form that makes its action, participants, and attributes explicit. We use the form \texttt{PREDICATE(Role: value, Role: value, \dots)}. For example, \texttt{PEEL(Agent: you, Object: onion(Size: medium, Quantity: 1))}. SRL representations preserve distinctions in object identity, quantity, destination, surface, and manner while reducing variation in surface form. These arguments often determine whether a step is correct. This structure is useful for CaptainCook4D, where near-duplicate steps can have different descriptions or IDs,\footnote{For example, \textit{Peel-Peel 1 garlic clove}, Step ID 14, and \textit{Peel-Peel 1 garlic cloves}, Step ID 200.} and for EgoOops, where long step descriptions are paired with short error cues.

We construct SRL representations with both a proprietary model, GPT-5.3~\citep{openai2025gpt5}, and an open-weight model, Qwen3.6-27B~\citep{qwen3.6-27b}. The two generators follow the same constrained predicate--argument format. We treat their outputs as two alternative SRL variants and compare how differences in role labeling and event decomposition affect downstream performance. Appendix~\ref{app:semrep} gives additional SRL details, including the prompt and generation settings for each model and systematic differences with examples.

\paragraph{Action--object representations.} We also derive an action--object abstraction that retains the main predicate and a flat set of salient entities, including tools and contacted objects. It removes quantities, attributes, manner, and the semantic relations that distinguish roles such as Destination and Surface. This is used as an ablation to test how much procedural detail can be removed before mistake recognition drops.

\paragraph{Video-to-text grounding with VLMs.}
For each segment, the VLM receives the clip and the candidate descriptions for its task. The candidates provide task-specific closed-set guidance. The VLM returns a structured output containing \texttt{predicted\_description}, \texttt{action\_name}, and \texttt{objects}. Raw-text runs encode \texttt{predicted\_description} directly. The action--object ablation uses the two auxiliary fields. Grounded SRL runs retrieve a cached representation keyed only by the normalized \texttt{predicted\_description}. Annotated Step IDs, canonical descriptions, and modified descriptions do not enter this lookup.

% \paragraph{Task interface.} Each procedure provides candidate steps, valid reference sequences, and correct executions. The candidate steps define the task-local state inventory. We estimate emission and transition parameters from correct executions. The reference sequences constrain local transitions and support edit alignment. PRISM applies the same grounding, representation, HMM estimation, decoding, and edit-alignment pipeline to every procedure.

\section{Experiments and results}
\label{sec:exp}

\begin{table*}[t]
\begin{center}
\small
\setlength{\tabcolsep}{1.2pt}
\begin{tabular}{lcccccccccccc}
\toprule
& \multicolumn{4}{c}{\bf EgoOops}
& \multicolumn{4}{c}{\bf EgoPER}
& \multicolumn{4}{c}{\bf CC4D} \\
\cmidrule(lr){2-5}\cmidrule(lr){6-9}\cmidrule(lr){10-13}
\bf Family
& \bf Seq & \bf Any-F1 & \bf Type-F1 & \bf AUC
& \bf Seq & \bf Any-F1 & \bf Type-F1 & \bf AUC
& \bf Seq & \bf Any-F1 & \bf Type-F1 & \bf AUC \\
\midrule
Gold text
& 0.713 & 0.480 & 0.446 & 0.756
& 0.903 & \textbf{0.741} & 0.673 & \textbf{0.918}
& 0.617 & \textbf{0.631} & 0.466 & \textbf{0.761} \\

Vision
& 0.328 & 0.454 & 0.439 & 0.523
& 0.372 & 0.519 & 0.527 & 0.580
& 0.159 & 0.556 & 0.464 & 0.478 \\

Gold + GPT-SRL
& \textbf{0.742} & \textbf{0.510} & \textbf{0.533} & \textbf{0.775}
& \textbf{0.913} & 0.736 & \textbf{0.675} & 0.908
& \textbf{0.638} & 0.629 & 0.495 & 0.757 \\

Qwen text
& 0.493 & 0.441 & 0.451 & 0.564
& 0.802 & 0.607 & 0.599 & 0.736
& 0.517 & 0.603 & 0.498 & 0.640 \\

Qwen + GPT-SRL
& 0.499 & 0.409 & 0.407 & 0.623
& 0.816 & 0.601 & 0.594 & 0.742
& 0.485 & 0.592 & 0.505 & 0.648 \\

Gemini text
& 0.557 & 0.408 & 0.361 & 0.610
& 0.845 & 0.642 & 0.589 & 0.748
& 0.577 & 0.603 & \textbf{0.516} & 0.656 \\

Gemini + GPT-SRL
& 0.560 & 0.353 & 0.314 & 0.611
& 0.854 & 0.631 & 0.595 & 0.749
& 0.554 & 0.595 & 0.513 & 0.669 \\
\bottomrule
\end{tabular}
\end{center}
\caption{Family-level leaderboard. Each row reports all metrics from one complete configuration. Encoder and HMM settings may differ across rows. Seq denotes sequence accuracy, and CC4D denotes CaptainCook4D. Appendix~\ref{app:family_configs} lists the corresponding configurations.}
\label{tab:main_results}
\end{table*}

\begin{table*}[t]
\begin{center}
\scriptsize
\setlength{\tabcolsep}{3pt}
\resizebox{\textwidth}{!}{%
\begin{tabular}{lcccc|cccc|cccc}
\toprule
& \multicolumn{4}{c|}{\bf Gold}
& \multicolumn{4}{c|}{\bf Qwen}
& \multicolumn{4}{c}{\bf Gemini} \\
\cmidrule(lr){2-5}
\cmidrule(lr){6-9}
\cmidrule(lr){10-13}
\bf Dataset
& \bf Raw & \bf GPT-SRL & $\boldsymbol{\Delta}$ & \bf Act.+Obj.
& \bf Raw & \bf GPT-SRL & $\boldsymbol{\Delta}$ & \bf Act.+Obj.
& \bf Raw & \bf GPT-SRL & $\boldsymbol{\Delta}$ & \bf Act.+Obj. \\
\midrule
EgoOops
& 0.400 & \textbf{0.434} & +0.034 & 0.318
& 0.321 & 0.356 & +0.036 & \textbf{0.444}
& 0.349 & 0.353 & +0.003 & \textbf{0.414} \\

EgoPER
& 0.681 & \textbf{0.725} & +0.044 & 0.540
& 0.578 & \textbf{0.601} & +0.024 & 0.550
& 0.602 & \textbf{0.631} & +0.030 & 0.552 \\

CaptainCook4D
& \textbf{0.629} & \textbf{0.629} & +0.000 & 0.571
& 0.587 & \textbf{0.592} & +0.005 & 0.566
& 0.595 & \textbf{0.596} & +0.001 & 0.570 \\
\bottomrule
\end{tabular}%
}
\end{center}
\caption{Any-F1 by text source and representation. Each Raw--GPT-SRL pair uses identical downstream settings, and $\Delta$ is GPT-SRL minus Raw. Act.+Obj. reports the best configuration within that representation family. Bold marks the highest score within each dataset--source group. Differences are computed before rounding.}
\label{tab:repr_grounding}
\end{table*}

\paragraph{Benchmarks and protocol.}
We adopt a common post-completion protocol across CaptainCook4D~\citep{peddi2024captaincook4ddatasetunderstandingerrors}, EgoPER~\citep{Lee_2024_CVPR}, and EgoOops~\citep{haneji2025egooopsdatasetmistakeaction}: the input is a fully observed, step-segmented procedural video, and the model must recover the executed step sequence and detect deviations from a valid reference procedure. We follow the text augmentation and cleanup protocol described in Section~\ref{sec:vid_to_txt}.

For CaptainCook4D and EgoPER, we use the benchmark-provided train, validation, and test splits and train each HMM on correct training videos. For Table~\ref{tab:main_results}, we select one complete configuration within each representation family by validation Any-F1; Type-F1, AUC, and Seq break ties in that order. Test data are used only for final reporting. Decision thresholds are fixed before test evaluation and reported in Appendix~\ref{app:hmm}. EgoOops has no official validation split. We train its HMMs on all 21 correct videos. For each representation family, we transfer the complete configuration with the highest mean validation Any-F1 across CaptainCook4D and EgoPER; Type-F1, AUC, and Seq again break ties. EgoOops error-video labels are not used for selection. The matched representation and decoder studies use the configurations reported in Table~\ref{tab:repr_grounding_configs}.

Under this step-level class imbalance ($20.8\%$--$39.2\%$ errors across datasets), accuracy can obscure poor error recall. We prioritize F1 and AUC metrics.

\paragraph{Models and representations.}
For visual step embeddings, we use frozen VideoMAE-v2~\citep{wang2023videomaev2} and V-JEPA 2~\citep{assran2025vjepa2selfsupervisedvideo} backbones. For text, we sweep frozen CLIP~\citep{radford2021learningtransferablevisualmodels}, all-MiniLM~\citep{wang2020minilmdeepselfattentiondistillation}, all-MPNet~\citep{song2020mpnetmaskedpermutedpretraining}, SimCSE~\citep{gao-etal-2021-simcse}, Instructor-XL~\citep{su2023embeddertaskinstructionfinetunedtext}, and OpenAI text-embedding-3-large~\citep{openai2024embeddingmodels}. For video-to-text grounding, we compare Gemini 3 Flash~\citep{google2025gemini3} and Qwen 3.5-27B~\citep{qwen3.5}. For \textit{Qwen + GPT-SRL} and \textit{Gemini + GPT-SRL}, we first generate grounded step descriptions with the VLM and then convert these generated descriptions into SRL representations\footnote{The main-paper SRL results use representations generated by GPT-5.3~\citep{openai2025gpt5}. We treat these as the primary SRL setting in the main comparisons, and report the corresponding Qwen3.6-27B~\citep{qwen3.6-27b} results in Appendix~\ref{app:additional_ablations}.} before text encoding and HMM inference. On the language side, we evaluate step text, semantic role representations, and action--object abstractions.

\begin{table*}[t]
\centering
\small
\setlength{\tabcolsep}{4pt}
\begin{tabular}{llcc|cc|cc}
\toprule
& & \multicolumn{2}{c|}{\bf Nearest candidate}
& \multicolumn{2}{c|}{\bf Greedy emission}
& \multicolumn{2}{c}{\bf Viterbi} \\
\cmidrule(lr){3-4}\cmidrule(lr){5-6}\cmidrule(lr){7-8}
\bf Dataset & \bf VLM
& \bf Raw & \bf SRL
& \bf Raw & \bf SRL
& \bf Raw & \bf SRL \\
\midrule
\multirow{2}{*}{EgoOops}
& Qwen   & 0.337 & 0.335 & \textbf{0.339} & 0.351 & 0.321 & \textbf{0.356} \\
& Gemini & 0.305 & 0.315 & 0.339 & 0.328 & \textbf{0.349} & \textbf{0.353} \\
\midrule
\multirow{2}{*}{EgoPER}
& Qwen   & 0.547 & 0.524 & \textbf{0.578} & \textbf{0.601}
                              & \textbf{0.578} & \textbf{0.601} \\
& Gemini & 0.571 & 0.557 & \textbf{0.602} & \textbf{0.631}
                              & \textbf{0.602} & \textbf{0.631} \\
\midrule
\multirow{2}{*}{CaptainCook4D}
& Qwen   & 0.567 & 0.559 & 0.581 & 0.578 & \textbf{0.587} & \textbf{0.592} \\
& Gemini & 0.538 & 0.528 & 0.564 & 0.561 & \textbf{0.595} & \textbf{0.596} \\
\bottomrule
\end{tabular}
\caption{Decoder controls under the matched Raw--SRL configurations. Nearest-candidate matching uses cosine similarity and no HMM. Greedy decoding uses learned emissions without start or transition probabilities. Viterbi uses the complete HMM. Bold marks the highest Any-F1 within each dataset--VLM--representation condition.}
\label{tab:decoder_controls}
\end{table*}

\begin{table*}[t]
\begin{center}
\small
\setlength{\tabcolsep}{4pt}
\begin{tabular}{lccccc}
\toprule
& \multicolumn{2}{c}{\bf EgoPER} & \multicolumn{3}{c}{\bf CaptainCook4D} \\
\cmidrule(lr){2-3} \cmidrule(lr){4-6}
\bf Method & \bf EDA & \bf AUC & \bf Precision & \bf EDA & \bf AUC \\
\midrule
% EgoPED~\citep{Lee_2024_CVPR}
% & 0.570 & 0.620
% & 0.565 & 0.698 & 0.549 \\

AMNAR
& 0.644 & 0.685
& 0.668 & 0.723 & 0.602 \\

AEM (pred. segments)
& 0.667 & 0.738
& 0.681 & 0.719 & 0.625 \\

AEM (GT segments)
& \textbf{0.747} & \textbf{0.867}
& -- & -- & -- \\

Ours (Qwen)
& 0.704 & 0.751
& 0.253 & 0.646 & 0.694 \\

Ours (Qwen + GPT-SRL)
& 0.699 & 0.753
& 0.785 & 0.841 & \textbf{0.900} \\

Ours (Gemini)
& 0.725 & 0.768
& 0.291 & 0.711 & 0.753 \\

Ours (Gemini + GPT-SRL)
& 0.713 & 0.762
& \textbf{0.809} & \textbf{0.843} & \textbf{0.900} \\

\bottomrule
\end{tabular}

\vspace{2pt}
\begin{minipage}{0.96\textwidth}
\footnotesize
\textit{Score sources.}
% EgoPED results on EgoPER come from \citet{Lee_2024_CVPR}. EgoPED CaptainCook4D results come from Table~3 of \citet{huang2025modelingmultiplenormalaction}, which evaluates EgoPED on that benchmark. 
AMNAR and AEM results come from their original papers~\citep{huang2025modelingmultiplenormalaction,guo2026proceduralmistakedetectionaction}. AEM (GT segments) is an inference-time ablation that replaces predicted segments with ground-truth segments.
\end{minipage}
\end{center}

\caption{System-level results under benchmark-specific protocols. PRISM uses benchmark-provided step boundaries. AEM results are shown for predicted and ground-truth (GT) segmentation. Baseline scores are taken from the original papers; all metrics in each PRISM row come from one configuration.}
\label{tab:prior_comparison_full}
\end{table*}

\paragraph{Metrics.}
We report four main metrics for the unified post-completion task.
\textbf{Any-F1} is our primary metric. It treats a segment as erroneous if it carries any of the four edit-based labels: Insertion, Deletion, Substitution, or Transposition.
\textbf{Type-F1} requires both the error location and its edit-based type to match the gold annotation.
\textbf{Seq} (Sequence accuracy) aligns the decoded and annotated executed Step-ID sequences with Needleman--Wunsch and reports the number of identical non-gap pairs divided by the number of evaluated segments.
\textbf{AUC} uses the negative emission score as a continuous anomaly score and measures threshold-free discrimination over observed segments. A gold Deletion is represented by an annotated position without an observed clip; it contributes to Any-F1 and Type-F1 but has no emission score and is excluded from AUC.

For comparison with prior work, we report Error Detection Accuracy (EDA) and area under the ROC curve (AUC) on EgoPER and CaptainCook4D and Precision only for CaptainCook4D, following the original EgoPED~\citep{Lee_2024_CVPR}, AMNAR~\citep{huang2025modelingmultiplenormalaction} and AEM~\citep{guo2026proceduralmistakedetectionaction}  evaluation protocols. We use these metrics only to compare with prior work.

For Tables~\ref{tab:main_results} and ~\ref{tab:prior_comparison_full}, all metrics within each row come from a single configuration. The family-level leaderboard uses the validation-based selection protocol described above. The matched Raw--SRL and decoder comparisons use fixed downstream configurations and perform no independent selection between conditions. For Table~\ref{tab:prior_comparison_full}, we select one configuration per family using validation EDA and report all benchmark-specific test metrics from that configuration.

\paragraph{Family-level results.}
Table~\ref{tab:main_results} compares representation families under the configuration-selection protocol described above. Gold-text families achieve the highest Any-F1 on each dataset. Grounded text-based variants outperform visual embeddings on EgoPER and CaptainCook4D, while vision performs better on EgoOops. This contrast highlights the importance of the available step descriptions. Within the separately selected grounded configurations, Raw achieves higher Any-F1 than SRL in all six pairs, although SRL improves Seq in four. Table~\ref{tab:repr_grounding} therefore examines representation changes under identical downstream settings.

\paragraph{Matched Raw--SRL comparison.}
Under identical downstream settings, GPT-SRL improves Any-F1 in all six grounded comparisons by 0.001–0.036. The largest gain occurs for Qwen on EgoOops. Gains on CaptainCook4D are small for both VLMs. Gold-description comparisons show improvements on EgoOops and EgoPER and a tie on CaptainCook4D. Together with Table~\ref{tab:main_results}, these results indicate that the benefit of SRL depends on the downstream configuration and selection criterion. Paired video-level bootstrap intervals for the Any-F1 difference exclude zero in three of the six grounded comparisons (Appendix~\ref{app:bootstrap}).

GPT-SRL also exceeds the independently selected Act.+Obj. configurations in seven of nine dataset–source settings; both exceptions occur with grounded EgoOops descriptions. This comparison complements the matched Raw–SRL analysis by showing the performance of a coarser representation.
Appendix~\ref{app:stratified_srl} reports effects by grounding stratum.

% \paragraph{One shared configuration.}
% Table~\ref{tab:universal} separates the best grounded pipeline on each dataset from one shared grounded configuration. The best per-dataset rows switch between grounded text without SRL and grounded text with GPT-SRL. The shared configuration uses Gemini predicted descriptions, GPT-SRL, Instructor-XL 128-d embeddings, no normalization, Gaussian emission, and difflib alignment. It is the strongest single shared setting by mean Any-F1 across all three datasets. Relative to per-dataset tuning, the Any-F1 loss is negligible on EgoOops and moderate on EgoPER and CaptainCook4D. 

\paragraph{Decoder controls.}
Table~\ref{tab:decoder_controls} separates independent candidate matching, the trained local decoder, and sequence decoding. Greedy emission decoding improves Any-F1 over nearest-candidate matching in all twelve conditions. Adding Viterbi improves seven conditions, matches four, and reduces one; it exceeds nearest-candidate matching in eleven conditions overall. EgoPER’s reported Any-F1 is unchanged by transition-aware decoding. The only decrease relative to greedy decoding occurs for Qwen Raw on EgoOops, where transition estimation relies on 21 correct videos across five tasks. These results show consistent benefits from the trained local decoder and dataset-dependent benefits from sequence decoding.

\paragraph{Comparison with prior work.}
Table~\ref{tab:prior_comparison_full} contextualizes PRISM against published systems under their respective protocols. On CaptainCook4D, the SRL configurations obtain EDA of 0.841–0.843 and AUC of 0.900. On EgoPER, AEM with ground-truth segmentation achieves higher EDA and AUC than PRISM. Differences in segmentation and configuration selection matter when interpreting these system-level results. Tables~\ref{tab:repr_grounding} and ~\ref{tab:decoder_controls} provide the controlled comparisons of representation and decoding within PRISM.
\section{Conclusion}
\label{sec:conclusion}

We introduced PRISM, a language-grounded framework for post-completion procedural mistake recognition that combines an HMM trained on correct executions with reference alignment and semantic checks. Across three benchmarks, the trained local decoder consistently improves over nearest-candidate matching, while sequence decoding provides additional, dataset-dependent gains. SRL improves Any-F1 in all six matched grounded comparisons, with benefits that depend on the dataset and downstream configuration. Our unified edit-based evaluation makes these effects comparable across benchmarks. Together, the results support language grounding as an effective interface for reasoning about procedural step content and order, making the underlying deviation explicit even when the failure appears only as a plate of overboiled zoodles.
\section*{Limitations}
\label{sec:limitations}

Our evaluation assumes benchmark-provided step boundaries, task-specific candidate descriptions, and valid reference procedures. Gold-text experiments bypass video grounding, whereas grounded experiments depend on the VLM preserving error-relevant details. The results therefore characterize mistake reasoning under these inputs. Deployment additionally requires reliable temporal segmentation and grounding, and adapting PRISM to a new procedure requires specifying its candidate steps and reference sequences and estimating its task-specific parameters.

Although the SRL generator comparison between GPT-5.3 and Qwen3.6-27B suggests that the pipeline is stable across SRL generators, the method still assumes consistent role normalization. Inconsistent semantic parses can propagate into HMM reasoning.

Our four label classes capture structural sequence edits, but they do not exhaust the full granularity of procedural mistake semantics. They align with recent unified mistake-taxonomy efforts, but for compatibility with current benchmarks we collapse Correction into Insertion and Wrong Execution into Substitution. This choice supports cross-dataset evaluation, at the cost of merging some semantically distinct cases. A natural next step is to refine the substitution class into full-step substitutions and role-level wrong executions, using SRL arguments to localize which part of a step changed and to support more fine-grained automated mistake attribution.
\section*{Ethical considerations}

This work uses existing benchmark datasets and does not involve new data collection, human-subject experiments, or deployment in real-world decision-making settings. Nevertheless, procedural mistake detection may raise ethical concerns in workplace, educational, or assistive settings, where video-based monitoring could be used for surveillance or overly punitive evaluation. Our method also depends on VLM-based grounding, so mistakes in generated step descriptions may propagate into downstream reasoning and lead to incorrect or misleading error labels. These systems should therefore be used only as decision-support tools under human oversight, rather than as standalone grounds for sanctions or safety-critical decisions. Any real deployment should additionally account for privacy, consent, secure data handling, and appropriate safeguards for affected users. Upon acceptance, we will release our code and scripts for generating SRL representations, in accordance with the licenses of the original
benchmarks.


\bibliography{custom}

@techreport{openai2025gpt5,
  title        = {{GPT-5} System Card},
  author       = {{OpenAI}},
  year         = {2025},
  month        = aug,
  institution  = {OpenAI},
  type         = {Technical report},
  url          = {https://cdn.openai.com/gpt-5-system-card.pdf},
  note         = {Version August 13, 2025}
}

@misc{google2025gemini3,
  title = {{Gemini} 3 {Flash}: Frontier Intelligence Built for Speed},
  author = {Doshi, Tulsee and the Gemini Team},
  year = {2025},
  month = dec,
  url = {https://blog.google/products/gemini/gemini-3-flash/},
  note = {Official announcement}
}

@misc{openai2024embeddingmodels,
  title = {New Embedding Models and {API} Updates},
  author = {{OpenAI}},
  year = {2024},
  month = jan,
  url = {https://openai.com/index/new-embedding-models-and-api-updates/},
  note = {Introduces text-embedding-3-large}
}

@misc{assran2025vjepa2selfsupervisedvideo,
      title={{V-JEPA 2}: Self-Supervised Video Models Enable Understanding, Prediction and Planning}, 
      author={Mido Assran and Adrien Bardes and David Fan and Quentin Garrido and Russell Howes and Mojtaba Komeili and Matthew Muckley and Ammar Rizvi and Claire Roberts and Koustuv Sinha and Artem Zholus and Sergio Arnaud and Abha Gejji and Ada Martin and Francois Robert Hogan and Daniel Dugas and Piotr Bojanowski and Vasil Khalidov and Patrick Labatut and Francisco Massa and Marc Szafraniec and Kapil Krishnakumar and Yong Li and Xiaodong Ma and Sarath Chandar and Franziska Meier and Yann LeCun and Michael Rabbat and Nicolas Ballas},
      year={2025},
      eprint={2506.09985},
      archivePrefix={arXiv},
      primaryClass={cs.AI},
      url={https://arxiv.org/abs/2506.09985}, 
}

@article{Byrne2006Task,
  author  = {Byrne, Michael D. and Davis, Elizabeth M.},
  title   = {Task Structure and Postcompletion Error in the Execution of a Routine Procedure},
  journal = {Human Factors},
  volume  = {48},
  number  = {4},
  pages   = {627--638},
  year    = {2006},
  doi     = {10.1518/001872006779166398},
  url     = {https://doi.org/10.1518/001872006779166398}
}

@article{Reason2000HumanEM,
  title={Human error: models and management.},
  author={James Reason},
  journal={BMJ},
  year={2000},
  volume={320 7237},
  pages={
          768-70
        },
  url={https://api.semanticscholar.org/CorpusID:261744128}
}

@article{understandingmodels,
author = {Garfield, Sara and Franklin, B.D.},
year = {2016},
month = {06},
pages = {361-364},
title = {Understanding models of error in clinical practice},
volume = {296},
journal = {Pharmaceutical Journal}
}

@book{Norman2002TheDO,
  author    = {Norman, Donald A.},
  title     = {The Design of Everyday Things},
  year      = {2002},
  publisher = {Basic Books},
  address   = {New York, N.Y.},
  edition   = {With a new introduction by the author},
  isbn      = {0465067107}
}

@misc{qian2022svipsequenceverificationprocedures,
      title={{SVIP}: Sequence VerIfication for Procedures in Videos}, 
      author={Yicheng Qian and Weixin Luo and Dongze Lian and Xu Tang and Peilin Zhao and Shenghua Gao},
      year={2022},
      eprint={2112.06447},
      archivePrefix={arXiv},
      primaryClass={cs.CV},
      url={https://arxiv.org/abs/2112.06447}, 
}

@misc{bacharidis2025visionbasedmistakeanalysisprocedural,
      title={Vision-Based Mistake Analysis in Procedural Activities: A Review of Advances and Challenges}, 
      author={Konstantinos Bacharidis and Antonis A. Argyros},
      year={2025},
      eprint={2510.19292},
      archivePrefix={arXiv},
      primaryClass={cs.CV},
      url={https://arxiv.org/abs/2510.19292}, 
}

@misc{peddi2024captaincook4ddatasetunderstandingerrors,
      title={{CaptainCook4D}: A Dataset for Understanding Errors in Procedural Activities}, 
      author={Rohith Peddi and Shivvrat Arya and Bharath Challa and Likhitha Pallapothula and Akshay Vyas and Bhavya Gouripeddi and Jikai Wang and Qifan Zhang and Vasundhara Komaragiri and Eric Ragan and Nicholas Ruozzi and Yu Xiang and Vibhav Gogate},
      year={2024},
      eprint={2312.14556},
      archivePrefix={arXiv},
      primaryClass={cs.CV},
      url={https://arxiv.org/abs/2312.14556}, 
}

@article{Sener2022Assembly101ALA,
  title={Assembly101: A Large-Scale Multi-View Video Dataset for Understanding Procedural Activities},
  author={Fadime Sener and Dibyadip Chatterjee and Daniel Shelepov and Kun He and Dipika Singhania and Robert Wang and Angela Yao},
  journal={2022 IEEE/CVF Conference on Computer Vision and Pattern Recognition (CVPR)},
  year={2022},
  pages={21064-21074},
  url={http://ieeexplore.ieee.org/stamp/stamp.jsp?tp=\&arnumber=9878834}
}

@article{Wang2023HoloAssistAEA,
  title={{HoloAssist}: an Egocentric Human Interaction Dataset for Interactive {AI} Assistants in the Real World},
  author={Xin Wang and Taein Kwon and Mahdi Rad and Bowen Pan and Ishani Chakraborty and Sean Andrist and D. Bohus and Ashley Feniello and Bugra Tekin and F. Frujeri and Neel Joshi and Marc Pollefeys},
  journal={2023 IEEE/CVF International Conference on Computer Vision (ICCV)},
  year={2023},
  pages={20213-20224},
  url={https://api.semanticscholar.org/CorpusId:263310718}
}

@article{Ding2023EveryMCA,
  title={Every Mistake Counts in Assembly},
  author={Guodong Ding and Fadime Sener and Shugao Ma and Angela Yao},
  journal={ArXiv},
  year={2023},
  volume={abs/2307.16453},
  url={https://api.semanticscholar.org/CorpusId:260334625}
}

@misc{haneji2025egooopsdatasetmistakeaction,
      title={{EgoOops}: A Dataset for Mistake Action Detection from Egocentric Videos Referring to Procedural Texts}, 
      author={Yuto Haneji and Taichi Nishimura and Hirotaka Kameko and Keisuke Shirai and Tomoya Yoshida and Keiya Kajimura and Koki Yamamoto and Taiyu Cui and Tomohiro Nishimoto and Shinsuke Mori},
      year={2025},
      eprint={2410.05343},
      archivePrefix={arXiv},
      primaryClass={cs.CV},
      url={https://arxiv.org/abs/2410.05343}, 
}

@InProceedings{Lee_2024_CVPR,
                    author    = {Lee, Shih-Po and Lu, Zijia and Zhang, Zekun and Hoai, Minh and Elhamifar, Ehsan},
                    title     = {Error Detection in Egocentric Procedural Task Videos},
                    booktitle = {Proceedings of the IEEE/CVF Conference on Computer Vision and Pattern Recognition (CVPR)},
                    month     = {June},
                    year      = {2024},
                    pages     = {18655-18666}
            }

@misc{schoonbeek2023industrealdatasetprocedurestep,
      title={{IndustReal}: A Dataset for Procedure Step Recognition Handling Execution Errors in Egocentric Videos in an Industrial-Like Setting}, 
      author={Tim J. Schoonbeek and Tim Houben and Hans Onvlee and Peter H. N. de With and Fons van der Sommen},
      year={2023},
      eprint={2310.17323},
      archivePrefix={arXiv},
      primaryClass={cs.CV},
      url={https://arxiv.org/abs/2310.17323}, 
}

@INPROCEEDINGS{9897369,
  author={Moriwaki, Kosuke and Nakano, Gaku and Inoshita, Tetsuo},
  booktitle={2022 IEEE International Conference on Image Processing (ICIP)}, 
  title={The {BRIO-TA} Dataset: Understanding Anomalous Assembly Process in Manufacturing}, 
  year={2022},
  volume={},
  number={},
  pages={1991-1995},
  doi={10.1109/ICIP46576.2022.9897369}}

@article{Jang2019EPICTentAE,
  title={{EPIC-Tent}: An Egocentric Video Dataset for Camping Tent Assembly},
  author={Youngkyoon Jang and Brian T. Sullivan and Casimir J. H. Ludwig and Iain D. Gilchrist and Dima Damen and W. Mayol-Cuevas},
  journal={2019 IEEE/CVF International Conference on Computer Vision Workshop (ICCVW)},
  year={2019},
  pages={4461-4469},
  url={https://api.semanticscholar.org/CorpusID:203325811}
}

@misc{huang2025modelingmultiplenormalaction,
      title={Modeling Multiple Normal Action Representations for Error Detection in Procedural Tasks}, 
      author={Wei-Jin Huang and Yuan-Ming Li and Zhi-Wei Xia and Yu-Ming Tang and Kun-Yu Lin and Jian-Fang Hu and Wei-Shi Zheng},
      year={2025},
      eprint={2503.22405},
      archivePrefix={arXiv},
      primaryClass={cs.CV},
      url={https://arxiv.org/abs/2503.22405}, 
}

@article{Flaborea2024PREGOOMA,
  title={{PREGO}: Online Mistake Detection in PRocedural EGOcentric Videos},
  author={Alessandro Flaborea and Guido Maria D'Amely di Melendugno and Leonardo Plini and Luca Scofano and Edoardo De Matteis and Antonino Furnari and G. Farinella and Fabio Galasso},
  journal={2024 IEEE/CVF Conference on Computer Vision and Pattern Recognition (CVPR)},
  year={2024},
  pages={18483-18492},
  url={https://api.semanticscholar.org/CorpusId:268856941}
}

@article{Plini2024TIPREGOCOA,
  title={{TI-PREGO}: Chain of Thought and In-Context Learning for Online Mistake Detection in PRocedural EGOcentric Videos},
  author={Leonardo Plini and Luca Scofano and Edoardo De Matteis and Guido Maria D'Amely di Melendugno and Alessandro Flaborea and Andrea Sanchietti and G. Farinella and Fabio Galasso and Antonino Furnari},
  journal={ArXiv},
  year={2024},
  volume={abs/2411.02570},
  url={https://api.semanticscholar.org/CorpusId:273820991}
}

@misc{seminara2025differentiabletaskgraphlearning,
      title={Differentiable Task Graph Learning: Procedural Activity Representation and Online Mistake Detection from Egocentric Videos}, 
      author={Luigi Seminara and Giovanni Maria Farinella and Antonino Furnari},
      year={2025},
      eprint={2406.01486},
      archivePrefix={arXiv},
      primaryClass={cs.CV},
      url={https://arxiv.org/abs/2406.01486}, 
}

@misc{narasimhan2023learningverificationtaskstructure,
      title={Learning and Verification of Task Structure in Instructional Videos}, 
      author={Medhini Narasimhan and Licheng Yu and Sean Bell and Ning Zhang and Trevor Darrell},
      year={2023},
      eprint={2303.13519},
      archivePrefix={arXiv},
      primaryClass={cs.CV},
      url={https://arxiv.org/abs/2303.13519}, 
}

@InProceedings{wang2023videomaev2,
    author    = {Wang, Limin and Huang, Bingkun and Zhao, Zhiyu and Tong, Zhan and He, Yinan and Wang, Yi and Wang, Yali and Qiao, Yu},
    title     = {{VideoMAE V2}: Scaling Video Masked Autoencoders With Dual Masking},
    booktitle = {Proceedings of the IEEE/CVF Conference on Computer Vision and Pattern Recognition (CVPR)},
    month     = {June},
    year      = {2023},
    pages     = {14549-14560}
}

@misc{mazzamuto2025gazingmisstepsleveragingeyegaze,
      title={Gazing Into Missteps: Leveraging Eye-Gaze for Unsupervised Mistake Detection in Egocentric Videos of Skilled Human Activities}, 
      author={Michele Mazzamuto and Antonino Furnari and Yoichi Sato and Giovanni Maria Farinella},
      year={2025},
      eprint={2406.08379},
      archivePrefix={arXiv},
      primaryClass={cs.CV},
      url={https://arxiv.org/abs/2406.08379}, 
}

@misc{storks2025transparentcoherentproceduralmistake,
      title={Transparent and Coherent Procedural Mistake Detection}, 
      author={Shane Storks and Itamar Bar-Yossef and Yayuan Li and Zheyuan Zhang and Jason J. Corso and Joyce Chai},
      year={2025},
      eprint={2412.11927},
      archivePrefix={arXiv},
      primaryClass={cs.AI},
      url={https://arxiv.org/abs/2412.11927}, 
}

@misc{ding2023temporalactionsegmentationanalysis,
      title={Temporal Action Segmentation: An Analysis of Modern Techniques}, 
      author={Guodong Ding and Fadime Sener and Angela Yao},
      year={2023},
      eprint={2210.10352},
      archivePrefix={arXiv},
      primaryClass={cs.CV},
      url={https://arxiv.org/abs/2210.10352}, 
}

@article{Ding2025TheJO,
  title={The Journey of Action Recognition},
  author={Xi Ding and Lei Wang},
  journal={Companion Proceedings of the ACM on Web Conference 2025},
  year={2025},
  url={https://api.semanticscholar.org/CorpusID:278822673}
}

@misc{lai2024humanactionanticipationsurvey,
      title={Human Action Anticipation: A Survey}, 
      author={Bolin Lai and Sam Toyer and Tushar Nagarajan and Rohit Girdhar and Shengxin Zha and James M. Rehg and Kris Kitani and Kristen Grauman and Ruta Desai and Miao Liu},
      year={2024},
      eprint={2410.14045},
      archivePrefix={arXiv},
      primaryClass={cs.CV},
      url={https://arxiv.org/abs/2410.14045}, 
}

@InProceedings{Lee_2025_ICCV,
    author    = {Lee, Shih-Po and Elhamifar, Ehsan},
    title     = {Error Recognition in Procedural Videos using Generalized Task Graph},
    booktitle = {Proceedings of the IEEE/CVF International Conference on Computer Vision (ICCV)},
    month     = {October},
    year      = {2025},
    pages     = {10009-10021}
}

@misc{radford2021learningtransferablevisualmodels,
      title={Learning Transferable Visual Models From Natural Language Supervision}, 
      author={Alec Radford and Jong Wook Kim and Chris Hallacy and Aditya Ramesh and Gabriel Goh and Sandhini Agarwal and Girish Sastry and Amanda Askell and Pamela Mishkin and Jack Clark and Gretchen Krueger and Ilya Sutskever},
      year={2021},
      eprint={2103.00020},
      archivePrefix={arXiv},
      primaryClass={cs.CV},
      url={https://arxiv.org/abs/2103.00020}, 
}

@misc{su2023embeddertaskinstructionfinetunedtext,
      title={One Embedder, Any Task: Instruction-Finetuned Text Embeddings}, 
      author={Hongjin Su and Weijia Shi and Jungo Kasai and Yizhong Wang and Yushi Hu and Mari Ostendorf and Wen-tau Yih and Noah A. Smith and Luke Zettlemoyer and Tao Yu},
      year={2023},
      eprint={2212.09741},
      archivePrefix={arXiv},
      primaryClass={cs.CL},
      url={https://arxiv.org/abs/2212.09741}, 
}

@inproceedings{gao-etal-2021-simcse,
    title = "{S}im{CSE}: Simple Contrastive Learning of Sentence Embeddings",
    author = "Gao, Tianyu  and
      Yao, Xingcheng  and
      Chen, Danqi",
    editor = "Moens, Marie-Francine  and
      Huang, Xuanjing  and
      Specia, Lucia  and
      Yih, Scott Wen-tau",
    booktitle = "Proceedings of the 2021 Conference on Empirical Methods in Natural Language Processing",
    month = nov,
    year = "2021",
    address = "Online and Punta Cana, Dominican Republic",
    publisher = "Association for Computational Linguistics",
    url = "https://aclanthology.org/2021.emnlp-main.552/",
    doi = "10.18653/v1/2021.emnlp-main.552",
    pages = "6894--6910"
}

@misc{wang2020minilmdeepselfattentiondistillation,
      title={{MiniLM}: Deep Self-Attention Distillation for Task-Agnostic Compression of Pre-Trained Transformers}, 
      author={Wenhui Wang and Furu Wei and Li Dong and Hangbo Bao and Nan Yang and Ming Zhou},
      year={2020},
      eprint={2002.10957},
      archivePrefix={arXiv},
      primaryClass={cs.CL},
      url={https://arxiv.org/abs/2002.10957}, 
}

@misc{song2020mpnetmaskedpermutedpretraining,
      title={{MPNet}: Masked and Permuted Pre-training for Language Understanding}, 
      author={Kaitao Song and Xu Tan and Tao Qin and Jianfeng Lu and Tie-Yan Liu},
      year={2020},
      eprint={2004.09297},
      archivePrefix={arXiv},
      primaryClass={cs.CL},
      url={https://arxiv.org/abs/2004.09297}, 
}

@article{gildea-jurafsky-2002-automatic,
    title = "Automatic Labeling of Semantic Roles",
    author = "Gildea, Daniel  and
      Jurafsky, Daniel",
    journal = "Computational Linguistics",
    volume = "28",
    number = "3",
    year = "2002",
    address = "Cambridge, MA",
    publisher = "MIT Press",
    url = "https://aclanthology.org/J02-3001/",
    doi = "10.1162/089120102760275983",
    pages = "245--288"
}

@article{palmer-etal-2005-proposition,
    title = "The {P}roposition {B}ank: An Annotated Corpus of Semantic Roles",
    author = "Palmer, Martha  and
      Gildea, Daniel  and
      Kingsbury, Paul",
    journal = "Computational Linguistics",
    volume = "31",
    number = "1",
    year = "2005",
    url = "https://aclanthology.org/J05-1004/",
    doi = "10.1162/0891201053630264",
    pages = "71--106"
}

@inproceedings{baker-etal-1998-berkeley-framenet,
    title = "The {B}erkeley {F}rame{N}et Project",
    author = "Baker, Collin F.  and
      Fillmore, Charles J.  and
      Lowe, John B.",
    booktitle = "36th Annual Meeting of the Association for Computational Linguistics and 17th International Conference on Computational Linguistics, Volume 1",
    month = aug,
    year = "1998",
    address = "Montreal, Quebec, Canada",
    publisher = "Association for Computational Linguistics",
    url = "https://aclanthology.org/P98-1013/",
    doi = "10.3115/980845.980860",
    pages = "86--90"
}

@inproceedings{batra-etal-2025-predicting,
    title = "Predicting Implicit Arguments in Procedural Video Instructions",
    author = "Batra, Anil  and
      Sevilla-Lara, Laura  and
      Rohrbach, Marcus  and
      Keller, Frank",
    editor = "Che, Wanxiang  and
      Nabende, Joyce  and
      Shutova, Ekaterina  and
      Pilehvar, Mohammad Taher",
    booktitle = "Proceedings of the 63rd Annual Meeting of the Association for Computational Linguistics (Volume 1: Long Papers)",
    month = jul,
    year = "2025",
    address = "Vienna, Austria",
    publisher = "Association for Computational Linguistics",
    url = "https://aclanthology.org/2025.acl-long.1467/",
    doi = "10.18653/v1/2025.acl-long.1467",
    pages = "30399--30419",
    ISBN = "979-8-89176-251-0"
}

@misc{li2025mistakeattributionfinegrainedmistake,
      title={Mistake Attribution: Fine-Grained Mistake Understanding in Egocentric Videos}, 
      author={Yayuan Li and Aadit Jain and Filippos Bellos and Jason J. Corso},
      year={2025},
      eprint={2511.20525},
      archivePrefix={arXiv},
      primaryClass={cs.CV}
}

@misc{sadhu2021visualsemanticrolelabeling,
      title={Visual Semantic Role Labeling for Video Understanding}, 
      author={Arka Sadhu and Tanmay Gupta and Mark Yatskar and Ram Nevatia and Aniruddha Kembhavi},
      year={2021},
      eprint={2104.00990},
      archivePrefix={arXiv},
      primaryClass={cs.CV},
      url={https://arxiv.org/abs/2104.00990}, 
}

@misc{guo2026proceduralmistakedetectionaction,
      title={Procedural Mistake Detection via Action Effect Modeling}, 
      author={Wenliang Guo and Yujiang Pu and Yu Kong},
      year={2026},
      eprint={2512.03474},
      archivePrefix={arXiv},
      primaryClass={cs.CV},
      url={https://arxiv.org/abs/2512.03474}, 
}

@misc{chen2025semanticrolelabelingsystematical,
      title={Semantic Role Labeling: A Systematical Survey}, 
      author={Huiyao Chen and Meishan Zhang and Jing Li and Min Zhang and Lilja Øvrelid and Jan Hajič and Hao Fei},
      year={2025},
      eprint={2502.08660},
      archivePrefix={arXiv},
      primaryClass={cs.CL},
      url={https://arxiv.org/abs/2502.08660}, 
}

@misc{qwen3.5,
    title  = {{Qwen3.5}: Towards Native Multimodal Agents},
    author = {{Qwen Team}},
    month  = {February},
    year   = {2026},
    url    = {https://qwen.ai/blog?id=qwen3.5}
}

@InProceedings{Loginova_2026_CVPR,
    author    = {Loginova, Olga and Keller, Frank},
    title     = {How to Correctly Make Mistakes: A Framework for Constructing and Benchmarking Mistake Aware Egocentric Procedural Videos},
    booktitle = {Proceedings of the IEEE/CVF Conference on Computer Vision and Pattern Recognition (CVPR) Workshops},
    month     = {June},
    year      = {2026},
    pages     = {8843-8853}
}

@misc{qwen3.6-27b,
    title  = {{Qwen3.6-27B}: Flagship-Level Coding in a {27B} Dense Model},
    author = {{Qwen Team}},
    month  = {April},
    year   = {2026},
    url    = {https://qwen.ai/blog?id=qwen3.6-27b}
}

@ARTICLE{1054010,
  author={Viterbi, A.},
  journal={IEEE Transactions on Information Theory}, 
  title={Error bounds for convolutional codes and an asymptotically optimum decoding algorithm}, 
  year={1967},
  volume={13},
  number={2},
  pages={260-269},
  doi={10.1109/TIT.1967.1054010}}

\appendix
\section{Dataset-specific annotation profiles}
\label{app:dataset_profiles}

The benchmarks differ in scale, domain, annotation style, and lexical variability. Table~\ref{tab:dataset_statistics} summarizes CaptainCook4D~\citep{peddi2024captaincook4ddatasetunderstandingerrors}, EgoPER~\citep{Lee_2024_CVPR}, and EgoOops~\citep{haneji2025egooopsdatasetmistakeaction}. CaptainCook4D spans 24 cooking tasks. EgoPER has a similar number of videos concentrated in five cooking tasks. EgoOops covers five non-cooking procedures. The following profiles describe the annotation differences that affect representation and evaluation.

\begin{center}
\small
\setlength{\tabcolsep}{3pt}
\begin{tabular}{lrrrr}
\toprule
\bf Dataset & \bf Videos & \bf Tasks & \bf Hours & \bf Step instances \\
\midrule
CaptainCook4D & 384 & 24 & 94.5 & 5.3K \\
EgoPER        & 386 & 5  & 28.0 & 5,369 \\
EgoOops       & 50  & 5  & 6.8  & 538 \\
\bottomrule
\end{tabular}
\captionsetup{type=table,hypcap=false}
\caption{Benchmark scale. Hours denote total video duration. The Step instances column reports annotated procedural units used for step-level training and evaluation. The benchmark papers use different names for these units.}
\label{tab:dataset_statistics}
\end{center}

\paragraph{CaptainCook4D.}
CaptainCook4D contains cooking procedures with relatively short step descriptions, but the surface forms are noisy.
The benchmark includes duplicated predicate prefixes such as \textit{Pour-Pour} and \textit{Peel-Peel}, which we remove during text cleanup because they do not add procedural meaning.
The dataset also contains near-duplicate step descriptions with different IDs, for example singular/plural variants such as \textit{Peel 1 garlic clove} and \textit{Peel 1 garlic cloves}.
Modified descriptions often encode quantity, duration, temperature, instrument, or destination changes inside the step arguments.
For example, \textit{Microwave the plate, covered, on high for 5 minutes instead of 1.5 minutes} differs from the canonical step mainly through duration. Similarly, the contrast between \textit{Cut the English muffin into two pieces with a knife} and \textit{Cut the English muffin into two pieces with hand} reflects a change in instrument, and the contrast between \textit{Top cup with 1 tablespoon of salsa} and \textit{Top cup with 1 teaspoon of salsa} reflects a change in quantity.
These cases illustrate why argument-level information remains important for mistake recognition. SRL preserves these distinctions in an explicit predicate--argument form.

\paragraph{EgoPER.}
EgoPER uses short and fairly regular cooking instructions in a visually consistent environment.
Its step descriptions are more atomic than in EgoOops and less lexically variable than in CaptainCook4D.
Many errors are local substitutions of action, object, or destination, such as changing the target surface or replacing the intended manipulation with a visibly different one. For example, \textit{Place tortilla on cutting board} becomes \textit{Place tortilla on table}, and \textit{Transfer water to kettle} becomes \textit{Spill out water}. Because text descriptions and their deviations are often short, explicit, and lexically close, text descriptions and SRL remain close in performance.
Action--object abstractions are still weaker because they remove distinctions that remain useful for mistake recognition, but the drop is more interpretable than in CaptainCook4D or EgoOops.

\paragraph{EgoOops.}
EgoOops has the most heterogeneous step language.
Some canonical steps are long and contain multiple subevents, for example \textit{Pour about 15mL of water into a cup, dip the tip of a red highlighter in the water and squeeze out a drop.}
Its error annotations are also unusual: the benchmark often provides a short error cue instead of a full observed step description. We therefore construct complete modified descriptions by concatenating the canonical step with the error cue.
For example, the canonical step \textit{Turn on the switch S2.} plus the error cue \textit{Blow up the propeller.} becomes \textit{Turn on the switch S2. Blow up the propeller.} Other cases modify object identity or orientation inside the step, for example \textit{Connect the battery box and switch S1 in series.} becoming \textit{Connect the battery box and switch S2 instead of switch S1 in series.} EgoOops errors may also introduce an extra event inside a step, for example \textit{Bridge two long thin cuboid blocks on the upright blocks from left to right in the order of yellow and red. Collapse blocks.} These patterns make the benchmark especially sensitive to role fillers, argument structure, and multi-event step descriptions. SRL makes these roles and appended deviations explicit.

\paragraph{Grounded descriptions.}
The same benchmark differences also affect video-to-text grounding.
Gemini usually produces gold-aligned step descriptions more accurately than Qwen, especially on CaptainCook4D.
On EgoPER, both VLMs are closer to the candidate-step vocabulary because the step language is shorter and more regular.
On EgoOops, grounded descriptions remain harder because the benchmark steps can be long, compositional, and error cues may add an extra event to an otherwise canonical step.

\begin{table*}[t]
\begin{center}
\small
\setlength{\tabcolsep}{3pt}
\begin{tabular}{@{}>{\raggedright\arraybackslash}p{0.14\textwidth}
                >{\raggedright\arraybackslash}p{0.64\textwidth}
                >{\raggedright\arraybackslash}p{0.14\textwidth}@{}}
\toprule
\textbf{Dataset} & \textbf{Native label / description} & \textbf{Unified label} \\
\midrule

\multirow{9}{*}{CC4D}
& Order Error & Transposition \\
& Missing Step & Deletion \\
& Preparation Error & Substitution \\
& Measurement Error & Substitution \\
& Timing Error & Substitution \\
& Technique Error & Substitution \\
& Temperature Error & Substitution \\
& Other & Substitution \\
& additional executed step outside the reference sequence & Insertion \\
\midrule

\multirow{5}{*}{EgoPER}
& step omission & Deletion \\
& addition & Insertion \\
& correction & Insertion \\
& modification & Substitution \\
& slip & Substitution \\
& step aligned as a deletion and insertion of the same Step ID
& Transposition \\
\midrule

\multirow{17}{*}{EgoOops}
& working with wrong objects & Substitution \\
& grasping wrong objects and releasing them without using & Insertion \\
& correction of mistake actions & Insertion \\
& unintended actions & Substitution \\
& working in the wrong way or moving & Substitution \\
& others: mix all the liquid and shine a black light on it & Insertion \\
& others: not able to fold because there are no incisions & Substitution \\
& others: can't fold cardboard & Substitution \\
& others: after turning on the switch s1, turn off it & Substitution \\
& others: at the same time, also turn on the switch s1 then turn off it & Substitution \\
& others: place two magnesium plates on the last step and put one back & Substitution \\
& others: pour much water but should pour about 15 ml / knock over an empty cup & Substitution \\
& others: pour powdered detergent into a cup of yellow liquid but should mix in an empty cup & Substitution \\
& others: put two drops on a copper plate & Substitution \\
& others: scatter pieces of zinc / put three zinc plates on the right column, but should on the center column & Substitution \\
& others: put two drops on the metal plate in the center of the microplate & Substitution \\
& others: place blocks from left to right in the order of red and yellow but should put them in the order of yellow and red & Substitution \\
\bottomrule
\end{tabular}
\end{center}
\caption{Mapping from benchmark-specific error labels to the unified edit-based taxonomy used in this work. For EgoOops, the rare \texttt{others} cases are resolved individually from the textual error descriptions. CC4D is CaptainCook4D.}
\label{tab:label_mapping}
\end{table*}

\section{Unified label mapping}
\label{app:label_mapping}

We map the benchmark-specific error annotations to a unified edit-based taxonomy with four labels: \emph{Insertion}, \emph{Deletion}, \emph{Substitution}, and \emph{Transposition}. Table~\ref{tab:label_mapping} summarizes the dataset-specific mappings. This mapping defines the gold targets for Type-F1 and converts native annotations into the four edit labels predicted after sequence alignment. It does not affect HMM training or Step-ID decoding. Because the benchmarks use different error inventories and annotation conventions, the mapping rules are dataset-specific.

When a task admits several valid references, the evaluator selects the reference closest to the decoded sequence. Structural targets are instantiated relative to that reference, as detailed in Appendix~\ref{app:per_error}.

\paragraph{CaptainCook4D.}
CaptainCook4D provides explicit categories for erroneous steps.
Additional executed actions that are not part of the reference sequence are treated as \emph{Insertion}.
In the original dataset annotations, omitted steps are represented with \texttt{start\_time = -1.0} and \texttt{end\_time = -1.0}; we keep this convention when constructing deletion labels.

\paragraph{EgoPER.}
EgoPER defines a psychologically motivated taxonomy of procedural errors and does not include a native \emph{Transposition} category. We derive transpositions by aligning the annotated executed Step-ID sequence with the valid reference sequence and identifying steps that occur in a different position.
As in CaptainCook4D, omitted steps are represented by missing aligned actions and receive \texttt{start\_time = -1.0} and \texttt{end\_time = -1.0} in the derived sequence-level representation.

\paragraph{EgoOops.}
EgoOops often encodes the deviation as a short textual cue attached to an otherwise canonical step and does not contain \emph{Deletion} or \emph{Transposition}.
The native category \textit{others} is rare and is resolved case by case from the textual error description.
Within this group, only \textit{mix all the liquid and shine a black light on it} is mapped to \emph{Insertion}; all remaining listed \textit{others} cases are mapped to \emph{Substitution}.

\begin{table*}[t]
\centering
\small
\setlength{\tabcolsep}{3pt}
\begin{tabular}{@{}>{\raggedright\arraybackslash}p{0.11\textwidth}
                >{\raggedright\arraybackslash}p{0.08\textwidth}
                >{\raggedright\arraybackslash}p{0.05\textwidth}
                >{\raggedright\arraybackslash}p{0.24\textwidth}
                >{\raggedright\arraybackslash}p{0.44\textwidth}@{}}
\toprule
\textbf{Dataset} & \textbf{Type} & \textbf{ID} & \textbf{Step description} & \textbf{SRL representation} \\
\midrule

\multirow{3}{*}{CC4D}
& modified & 20 &
Microwave the plate, covered, on high for 5 minutes instead of 1.5 minutes &
\texttt{MICROWAVE(Agent: you, Object: plate(State: covered), Setting: high, Duration: minute(Quantity: 5), Manner: INSTEAD\_OF(Duration: minute(Quantity: 1.5)))} \\

& modified & 28 &
Measure 1/8 teaspoon of salt using dirty measuring spoon and add it to the mug &
\texttt{ADD(Agent: you, Object: salt(Measurement: teaspoon(Quantity: 1/8)), Destination: mug, Instrument: dirty(measuring\_spoon), Manner: after(MEASURE(Agent: you, Object: salt(Measurement: teaspoon(Quantity: 1/8)), Instrument: dirty(measuring\_spoon))))} \\

& canonical & 54 &
Put all the Vegetables in a microwave-safe bowl &
\texttt{PUT(Agent: you, Object: vegetables(Quantity: all), Destination: bowl(Property: microwave\_safe))} \\
\midrule

\multirow{2}{*}{EgoOops}
& canonical & 211 &
Put chopsticks through the cutout in the lid to widen the hole. &
\texttt{PUT(Agent: you, Object: chopsticks, Path: through(cutout(Location: in(lid))), Purpose: WIDEN(Agent: you, Object: hole))} \\

& modified & 10 &
Pour about 15mL of water into a cup, dip the tip of a red highlighter in the water and squeeze out a drop. Knock over an empty cup. &
\texttt{POUR(Agent: you, Object: water(Measurement: ml(Quantity: 15, Approx: about)), Destination: into(cup)); DIP(Agent: you, Object: tip(of(red\_highlighter)), Destination: in(water(Location: in(cup)))); SQUEEZE\_OUT(Agent: you, Object: drop); KNOCK\_OVER(Agent: you, Object: empty\_cup)} \\
\midrule

\multirow{3}{*}{EgoPER}
& canonical & 57 &
Put banana slices on tortilla &
\texttt{PUT(Agent: you, Object: banana\_slices, Destination: on(tortilla))} \\

& canonical & 37 &
Microwave for X seconds &
\texttt{MICROWAVE(Agent: you, Temporal: for(second(Quantity: x)))} \\

& canonical & 18 &
Drop tea bag &
\texttt{DROP(Agent: you, Object: tea\_bag)} \\
\bottomrule
\end{tabular}
\caption{Representative SRL examples from the three benchmarks. We include both canonical and modified step descriptions to show how the representation handles quantities, instruments, destinations, purposes, and multi-event modified steps. CC4D is CaptainCook4D.}
\label{tab:srl_examples}
\end{table*}

\begin{landscape}
\begin{table}[p]
  \centering
  \small
  \setlength{\tabcolsep}{2pt}
  \renewcommand{\arraystretch}{1.0}
  \begin{tabularx}{\textheight}{@{}>{\raggedright\arraybackslash}p{0.14\textheight}
                                >{\centering\arraybackslash}p{0.055\textheight}
                                >{\centering\arraybackslash}p{0.065\textheight}
                                YYY@{}}
    \toprule
    \textbf{Hyperparameter} & \textbf{Sym.} & \textbf{Tags} &
    \textbf{Meaning} &
    \textbf{Value effect} &
    \textbf{Default settings \& rationale} \\
    \midrule

    Frames per step & $n$ & t/i &
    \emph{Number of frames} per step averaged into the image feature. &
    High: stronger denoising and object coverage, but more IO/latency and possible over-smoothing on very short steps; low: faster and crisper, but noisier and more sensitive to accidental frames. &
    \textbf{3}: a good robustness--latency trade-off; using more frames may introduce irrelevant content in short steps but can help with long, stable steps. \\

    PCA dimension & $d$ & t/i &
    Output dimensionality after PCA (fit on train; applied at inference) before Gaussian emissions. &
    High: preserves fine semantics, but makes covariance estimation harder and can overfit; low: better conditioning and sample-efficiency, but too low drops discriminative cues. &
    \textbf{128}: stable Ledoit–Wolf covariances with modest memory/compute; increase with ample per-class data. \\

    Laplace smoothing & $\beta$ & t &
    Add-$\beta$ smoothing for transition $\mA$ and start prior $\boldsymbol{\pi}$ (counts+$\beta$, renormalize). &
    High: flatter $\mA,\boldsymbol{\pi}$, safer on rare or omitted transitions, and less brittle; low: sharper priors that mirror train paths, but more risk of overfitting ordering quirks. &
    \textbf{3.0}: applied before reference-based transition masking. \\

    Banding (local moves) & band & t &
    For each valid reference sequence, let $i$ and $j$ be the positions of two states in that sequence. Mask $A_{ss'}$ unless $|i{-}j|\leq1$ in at least one valid reference, then renormalize each row; encodes locally ordered procedures. &
    On: restricts non-local transitions and produces more locally ordered paths; off: allows global jumps for branched workflows, but raises transposition false positives. &
    \textbf{On} with $|i{-}j|\le1$: disabling recommended only if tasks truly need non-local transitions. \\

    Jitter probability & $p_{\text{jit}}$ & t &
    Probability to add Gaussian noise to a training vector before PCA. &
    High: more invariance to ambience and lighting, but too much blurs class boundaries; low: crisper clusters, but more brittle to domain shift. &
    \textbf{0.15}: mild diversity without collapsing adjacent steps; co-tuned with $\sigma$. \\
    \bottomrule
  \end{tabularx}
   \caption{Core HMM hyperparameters for sequence modeling and training-time regularization. Tags indicate whether a hyperparameter is used during training (\textbf{t}) or inference (\textbf{i}).}
  \label{tab:hmm_hparams}
\end{table}
\end{landscape}

\begin{landscape}
\begin{table}[p]
  \centering
  \small
  \setlength{\tabcolsep}{2pt}
  \renewcommand{\arraystretch}{1.0}
  \begin{tabularx}{\textheight}{@{}>{\raggedright\arraybackslash}p{0.14\textheight}
                                >{\centering\arraybackslash}p{0.055\textheight}
                                >{\centering\arraybackslash}p{0.065\textheight}
                                YYY@{}}
    \toprule
    \textbf{Hyperparameter} & \textbf{Sym.} & \textbf{Tags} &
    \textbf{Meaning} &
    \textbf{Value effect} &
    \textbf{Default settings \& rationale} \\
    \midrule

    Jitter std & $\sigma$ & t &
    Noise scale in $\mathcal{N}(0,\sigma^2 I)$ before PCA. &
    High: more robust, but can distort likelihoods and increase class overlap; low: minimal regularization. &
    \textbf{0.3} in configurations that use jitter. \\

    Per-modality normalization & l2/None & t/i &
    L2-normalize each modality before fusion and HMM; stabilizes scales and PCA. &
    l2: avoids dominance by high-variance channels and yields steadier thresholds; none: preserves raw magnitudes, but can skew fusion and Gaussian likelihoods. &
    \textbf{l2}: matches CLIP best practice; lets $w$ modulate content, not scale. \\

    Emission family & Gauss or Cosine & i &
    Observation model at inference: Gaussian log-likelihoods vs.\ cosine similarity (with temperature). &
    Gauss: calibrated log-likelihoods and good use of PCA plus Ledoit--Wolf; cosine: magnitude-invariant, but needs a temperature and may require transition rebalancing. &
    \textbf{Gauss} default; switch to \textbf{Cosine} for strongly angular features or when skipping PCA. \\

    Cosine temperature & $T$ & i &
    Scale for cosine emissions ($E=T\!\cdot\!\cos$); controls emission/transition balance. &
    High: more emission-dominant, peak-following paths; low: more transition-dominant, smoother paths. &
    \textbf{N/A} for Gauss. For Cosine, tuned $T$ to match the dynamic range of log-transitions. \\

    Emission scaling & $\alpha$ & i &
    Global multiplier $\tilde{\mE}=\alpha \mE$ to retune emission vs.\ transition influence. &
    High: more emission-driven Viterbi, hence more sensitive to per-step scores; low: more transition-driven, smoother, and more prior-driven. &
    \textbf{1.0}: adjustment needed only if domain shift skews score ranges. \\

    Semantic substitution threshold & $\tau_{\mathrm{sem}}$ & i &
    Cosine-similarity threshold between an observation and the mean embedding of its decoded state. A lower score produces a semantic Substitution when structural alignment retains the same Step ID. &
    High: detects subtler argument-level changes but may increase false positives; low: is more conservative but may miss modified executions. &
    \textbf{0.85} in all reported experiments. \\
    \bottomrule
  \end{tabularx}
   \caption{Core HMM hyperparameters for normalization and emission control. Tags indicate whether a hyperparameter is used during training (\textbf{t}) or inference (\textbf{i}).}
  \label{tab:hmm_hparams_emission}
\end{table}
\end{landscape}

\begin{landscape}
\begin{table}[p]
\centering
\small
\setlength{\tabcolsep}{3pt}
\renewcommand{\arraystretch}{1.08}

\begin{tabularx}{\linewidth}{
@{}
>{\raggedright\arraybackslash}p{0.15\linewidth}
>{\centering\arraybackslash}p{0.055\linewidth}
>{\centering\arraybackslash}p{0.05\linewidth}
X X X
@{}
}
% \begin{landscape}
% \begin{table}[p]
%   \centering
%   \small
%   \setlength{\tabcolsep}{3pt}
%   \renewcommand{\arraystretch}{1.06}
%   \begin{tabularx}{\textheight}{@{}>{\raggedright\arraybackslash}p{0.16\textheight}
%                                 >{\centering\arraybackslash}p{0.07\textheight}
%                                 >{\centering\arraybackslash}p{0.08\textheight}
%                                 YYY@{}}
    \toprule
    \textbf{Hyperparameter} & \textbf{Sym.} & \textbf{Tags} &
    \textbf{Meaning} &
    \textbf{Value effect} &
    \textbf{Default settings \& rationale} \\
    \midrule

    Emission anomaly threshold & $\tau_{\mathrm{em}}$ & i &
    Flag if the selected-state emission score is below $\tau_{\mathrm{em}}$. &
    Higher thresholds flag more observations; lower thresholds are more conservative. &
    $-10$ for Gaussian emissions and $0$ for cosine emissions. \\

    Transition anomaly threshold & $\tau_{\mathrm{tr}}$ & i &
    Flag if $\log A_{S_{t-1},S_t}<\tau_{\mathrm{tr}}$. &
    Higher thresholds flag more transitions; lower thresholds are more conservative. &
    $-23.5$ for CaptainCook4D and $-5$ for EgoPER and EgoOops. \\

    Alignment mode & NW or difflib & i &
    Align prediction to a canonical variant to label Deletion, Insertion, Substitution, or Transposition. &
    NW: stable edit metric (use sub=2, gap=1 for reports). \newline
    difflib: forgiving for narratives and qualitative inspection. &
    \textbf{difflib} is our default in the main experiments because it performs well across grounded settings; \textbf{NW} remains competitive and is preferred in some paper-facing or family-specific comparisons. \\

    Global fallback HMM & task$=-1$ & i &
    Task-agnostic model pooled over all tasks and available when task-specific parameters are missing. &
    Provides an implementation fallback when no task-specific model is available. &
    Implemented but not invoked in the reported experiments.\\

    Vision weight (mm) & $w$ & mm (t/i) &
    Fusion in joint space: $\bm{x}=w\,\bm{v}_{img}+(1-w)\,\bm{v}_{txt}$ (or scaled-concat in disjoint space). &
    High: vision-led. \newline
    Low: text-led; better when phrasing disambiguates or clarifies visuals. &
    \textbf{Task-dependent}. Optimal $w{\approx}0.2$–$0.4$: for subtle manipulations; increase when distinctive visuals drive the step. \\

    Use superstructure (mm) & on/off & mm (t/i) &
    Add structured phrases (objects/actions) as text before fusion. &
    On: boosts separability for visually similar steps; sensitive to noisy phrases. \newline
    Off: simpler; may blur key discriminators. &
    \textbf{Off} for clean baselines; \textbf{On} in V\,+\,AO style variants with reliable extractors and prompt filtering. \\

    Superstructure top-$k$ & $k$ & mm (t/i) &
    Keep top-$k$ phrases by confidence to control context breadth. &
    High: broader context, more recall but more distractors. \newline
    Low: focused cues; risk missing secondary hints. &
    \textbf{$k=3$–$5$}: small slate gives gains without flooding the encoder. \\

    Superstructure threshold & $\tau$ & mm (t/i) &
    Minimum confidence for phrases included in superstructure. &
    High: precise/sparse cues. \newline
    Low: recall-oriented; may add noise and hurt calibration. &
    \textbf{Moderate $\tau$}: to be tuned jointly with $k$ for coverage vs.\ noise. \\
    \bottomrule
  \end{tabularx}
   \caption{Additional HMM hyperparameters for anomaly scoring, alignment, and multimodal variants. Tags indicate whether a hyperparameter is used during training (\textbf{t}), inference (\textbf{i}), or in a multimodal setting (\textbf{mm}); entries like \textbf{mm (t/i)} denote multimodal settings with separate training/inference use.}
  \label{tab:hmm_hparams_mm}
\end{table}
\end{landscape}

\section{Semantic representation construction}
\label{app:semrep}

Each step is mapped to an SRL representation of the form
\texttt{PREDICATE(Role: value, Role: value, ...)}.
The representation makes the main event, its participants, and key attributes explicit.
We use nested structures when the step meaning requires internal attachment, scope, or composition.
Table~\ref{tab:srl_examples} gives examples from all three benchmarks.

We generate SRL representations offline with two LLM generators: GPT-5.3~\citep{openai2025gpt5} and Qwen3.6-27B~\citep{qwen3.6-27b}. Both generators use the same constrained few-shot prompt and the same output fields. This keeps the target SRL interface fixed across models. The model-specific differences are in decoding and output enforcement: GPT-5.3 is queried with strict JSON-schema decoding at temperature $0.1$, while Qwen3.6-27B is run with thinking disabled, temperature $0.0$, top-$p=0.8$, and top-$k=20$, after which we parse the returned JSON object. 

The generation process covers three vocabularies:
(1) step descriptions from correct videos,
(2) modified descriptions for erroneous steps, and
(3) VLM-predicted descriptions that are not already covered by the first two sets.

Both generators use the same predicate--argument interface, but they produce different formalisms. For example, GPT-5.3 tends to preserve the surface predicate, while Qwen3.6-27B more often rewrites the instruction into a normalized action representation.

The two generators also differ in role inventory: GPT-5.3 uses 98 distinct role labels, while Qwen3.6-27B uses 71. Several GPT-5.3 labels correspond to distinctions that Qwen3.6-27B encodes with more general roles or nested modifiers.

For example, GPT-5.3 sometimes uses \texttt{Setting}, \texttt{Surface}, \texttt{Position}, and \texttt{Container}, while Qwen3.6-27B usually uses \texttt{Location}, \texttt{Destination}, or nested entity structure.
Similarly, GPT-5.3 separates labels such as \texttt{Duration}, \texttt{Range}, and \texttt{Approx}; Qwen3.6-27B usually places these distinctions inside \texttt{Temporal} or \texttt{Measurement}.
Quantity-related labels such as \texttt{Fraction}, \texttt{Portion}, and \texttt{Amount} are also more often collapsed into \texttt{Quantity} or \texttt{Measurement} in Qwen3.6-27B. Table~\ref{tab:srl_role_inventory_examples} gives more representative examples.

\begin{table*}[t]
\centering
\small
\setlength{\tabcolsep}{3pt}
\renewcommand{\arraystretch}{1.08}
\begin{tabular}{p{0.16\linewidth} p{0.27\linewidth} p{0.25\linewidth} p{0.25\linewidth}}
\toprule
\textbf{GPT-5.3 label} & \textbf{Step description} & \textbf{GPT-5.3} & \textbf{Qwen3.6-27B} \\
\midrule

Setting &
Microwave the ramekin cup uncovered on high for 10 seconds &
\texttt{Setting: high} &
\texttt{Manner: uncovered \& on(high)} \\

Surface &
Brush 2 slices of baguette with olive oil on both sides &
\texttt{Surface: both\_sides} &
\texttt{Location: on(both\_sides)} \\

Position &
Put three copper plates on the left column of the microplate &
\texttt{column(of(microplate), Position: left)} &
\shortstack[l]{\texttt{left\_column(}\\ \texttt{of(microplate))}} \\

Container &
In the mug, whisk two eggs &
\texttt{eggs(Quantity: 2, Container: mug)} &
\texttt{eggs(Quantity: 2), Location: in(mug)} \\

Duration &
Microwave the plate, uncovered, on high for 5 minutes &
\texttt{Duration: minute(Quantity: 5)} &
\texttt{Temporal: for(minute(Quantity: 5))} \\

Range &
Continue to microwave for 15--30 more seconds &
\texttt{seconds(Range: 15-30)} &
\texttt{seconds(Quantity: 15-30)} \\

Approx &
Slice one tomato into about 1/2 inch thick slices &
\texttt{Thickness: inch(Quantity: 0.5, Approx: about)} &
\texttt{Thickness: about(1/2(inch))} \\

Portion &
Slice 1/4 medium onion into pieces &
\texttt{Portion: quarter(Quantity: 1)} &
\texttt{Quantity: 1/4} \\

Fraction &
Add 1/4 teaspoon cinnamon to the mug &
\texttt{teaspoon(Fraction: 1/4)} &
\texttt{teaspoon(Quantity: 1/4)} \\

Amount &
Pour a small amount of milk into the filter &
\texttt{Amount: small} &
\texttt{Quantity: small\_amount} \\

Alternative\_to &
Squeeze out a drop of a green highlighter into a cup and pour much water instead of about 15mL of water &
\texttt{Alternative\_to: water(Measurement: ml(Quantity: 15, Approx: about))} &
\shortstack[l]{\texttt{Manner: instead\_of(}\\ \texttt{about(15mL(of(water))))}} \\

Contrast &
Add tomatoes to the pan, but splash some tomatoes outside &
\texttt{Contrast: BUT(SPLASH(...))} &
\texttt{ADD(...) \& SPLASH(...)} \\

\bottomrule
\end{tabular}
\caption{Examples of role-inventory differences between GPT-5.3 and Qwen3.6-27B. GPT-5.3 uses some low-frequency labels that Qwen3.6-27B often represents with broader roles, nested modifiers, or explicit event composition.}
\label{tab:srl_role_inventory_examples}
\end{table*}

Manual inspection calibrates the prompt and role schema. We inspect preliminary outputs for role consistency, attachment, lexical form, and coordination scope. Recurring issues lead to a schema clarification or an additional few-shot example. This iterative process produced the final nine-example prompt.

After the final inspection, each generator regenerates the complete dictionaries automatically. No roles or argument structures in the final dictionaries are edited manually. Scripted checks verify valid JSON, required fields, and balanced delimiters.

Prompt calibration uses ontology descriptions without access to split membership, mistake labels, HMM predictions, or evaluation results. Each unique description is converted once, independent of its video occurrence or dataset split.

We apply the same calibrated prompt to all unique descriptions across the three datasets.

\begin{table*}[t]
\centering
\scriptsize
\setlength{\tabcolsep}{3pt}
\renewcommand{\arraystretch}{1.08}
\begin{tabular}{p{0.17\linewidth} p{0.23\linewidth} p{0.27\linewidth} p{0.27\linewidth}}
\toprule
\textbf{Difference} & \textbf{Step description} & \textbf{GPT-5.3} & \textbf{Qwen3.6-27B} \\
\midrule

Predicate choice &
Use a butter knife to scoop nut butter from the jar; Use scissors to cut the bag in half &
\texttt{USE(Object: butter\_knife, Purpose: SCOOP(Object: nut\_butter, Origin: jar))}; \texttt{USE(Object: scissors, Purpose: CUT(Object: bag, Result: in\_half))} &
\texttt{SCOOP(Object: nut\_butter, Instrument: butter\_knife, Origin: jar)}; \texttt{CUT(Object: bag, Instrument: scissors, Result: in(half))} \\

Alternative action &
Chop or grate the cucumber &
\texttt{CHOP\_OR\_GRATE(Object: cucumber)} &
\texttt{CHOP(Object: cucumber) | GRATE(Object: cucumber)} \\

Event decomposition &
Measure and add 2 tbsp water to the bowl &
\texttt{ADD(Object: water(Measurement: tablespoon(Quantity: 2)), Destination: bowl, Manner: MEASURED)} &
\texttt{MEASURE(Object: water(Measurement: tablespoon(Quantity: 2))) \& ADD(Object: water(...), Destination: to(bowl))} \\

Predicate normalization &
Slice mushrooms &
\texttt{SLICE(Object: mushrooms)} &
\texttt{CUT(Object: mushrooms, Result: slices)} \\

State change &
Return to low heat &
\texttt{RETURN(Destination: to(low\_heat))} &
\texttt{SET(Object: heat, Result: low)} \\

Error-event scope &
Spill while cracking an egg in the bowl &
\texttt{SPILL(Temporal: WHILE(CRACK(Object: egg, Destination: in(bowl))))} &
\texttt{CRACK(Object: egg, Destination: bowl, Manner: spill)} \\

Negation &
Can't fold along the incisions of the cardboard &
\texttt{FOLD(Object: cardboard, Path: along(incisions(of(cardboard))), Polarity: negative, Ability: can\_not)} &
\texttt{NEG(CAN(Action: FOLD(Object: cardboard, Path: along(incisions(of(cardboard))))))} \\

Incorporated argument &
Pepper to taste &
\texttt{SEASON(Instrument: pepper, Manner: to\_taste)} &
\texttt{ADD(Object: pepper(Measurement: to\_taste))} \\

Coordination &
Extract and add contents of an egg to a mug &
\texttt{ADD(Object: contents(of(egg)), Destination: mug, Purpose: EXTRACT(Object: contents(of(egg))))} &
\texttt{EXTRACT(Object: contents(of(egg))) \& ADD(Object: contents(of(egg)), Destination: to(mug))} \\

\bottomrule
\end{tabular}
\caption{Differences between SRL representations generated by GPT-5.3 and Qwen3.6-27B. The two models use the same predicate--argument syntax but differ in predicate choice, role assignment, event decomposition, and scope representation. Examples are shortened where needed.}
\label{tab:srl_generator_comparison}
\end{table*}

Table~\ref{tab:srl_prompt_schema} summarizes the core prompt design and output schema.

\begin{table*}[t]
\centering
\small
\setlength{\tabcolsep}{4pt}
\begin{tabular}{p{0.20\linewidth} p{0.73\linewidth}}
\toprule
\textbf{Component} & \textbf{Content} \\
\midrule

System instruction &
You are a linguistic semantic analyzer. For each procedural step description, generate the semantic representation in roles. \\

Prompt guidance &
Predicates are uppercase. Concrete manipulated entities are usually encoded under \texttt{Object}. \texttt{Agent: you} is used unless another agent is explicitly specified. Entity names are normalized as lowercase\_with\_underscores. Nested structures are allowed when needed, e.g., \texttt{tip(of(red\_highlighter))} or \texttt{water(Location: in(cup))}. The prompt also gives typical role names such as \texttt{Object}, \texttt{Coobject}, \texttt{Destination}, \texttt{Location}, \texttt{Instrument}, \texttt{Manner}, \texttt{Purpose}, \texttt{Temporal}, \texttt{Origin}, and \texttt{Result}. \\

Selected few-shot examples &
\texttt{INSERT(Agent: you, Object: test\_swab, Destination: into(nostril(of(her))))}; \newline
\texttt{ADD(Agent: you, Object: coffee\_grounds, Origin: bowl, Destination: filter(Location: in(french\_press)))}; \newline
\texttt{ADD(Agent: you, Object: cut(onions), Coobject: egg(Location: in(mixing\_bowl)))}; \newline
\texttt{TOP(Agent: you, Object: cup, Coobject: with(salsa(Measurement: tablespoon(Quantity: 1))))}; \newline
\texttt{POUR(Agent: you, Object: egg(Quantity: 1), Destination: into(ramekin\_cup))}. \\

Output schema &
Strict JSON with one item per step, containing the fields \texttt{id}, \texttt{step\_description}, and \texttt{semantic\_representation}. \\

Prompt calibration and checks &
Iterative manual inspection calibrates the role schema and nine-example prompt. Each generator then regenerates the complete dictionaries automatically. Scripted checks verify valid JSON, required fields, and balanced delimiters. No final role or argument structure is manually edited. \\

Covered vocabularies &
Gold canonical descriptions, modified descriptions, and VLM-predicted descriptions not covered by the first two vocabularies. Each unique description is processed once. At grounded inference, the normalized VLM-generated description is the only SRL lookup key. \\

\bottomrule
\end{tabular}
\caption{Core design of the prompt and schema used to generate SRL representations.}
\label{tab:srl_prompt_schema}
\end{table*}

\section{HMM hyperparameters and computational details}
\label{app:hmm}

Tables~\ref{tab:hmm_hparams}--\ref{tab:hmm_hparams_mm} define the supported HMM parameters. Section~\ref{app:configs} reports the settings used in each experiment.

\paragraph{Computational details.}
Experiments were run on an institutional compute node with one NVIDIA A100 (80\,GB) GPU and an AMD EPYC 7282 CPU (8 cores and
56\,GB RAM allocated per job). HMM training and inference are
CPU-bound, and a per-task HMM trains in seconds; the cost of the
experiments comes from configuration sweeps. The complete multimodal
sweeps on CaptainCook4D, the largest benchmark, consumed
${\approx}57$ node-hours (${\approx}454$ CPU-core-hours). GPU compute is required for the frozen encoders and the locally hosted open-weight models: video-to-text grounding with Qwen~3.5-27B took roughly 24~GPU-hours on the largest benchmark and less on the other two, and SRL generation with Qwen3.6-27B for all three benchmarks took 2.65~GPU-hours (batch size~1, reasoning mode disabled). Across all benchmarks, representation families, and configuration sweeps, the reported experiments consumed several hundred node-hours in total. Gemini~3~Flash and GPT-5.3 were accessed through their public APIs. We do not train or fine-tune any neural model; parameter counts of the open models are given in the cited papers and model names, and those of the proprietary models are not publicly disclosed.

\section{Reported configurations}
\label{app:configs}

This section reports the configurations used for (1) the family-level leaderboard in Table~\ref{tab:main_results}, (2) the matched representation and decoder studies in Tables~\ref{tab:repr_grounding} and~\ref{tab:decoder_controls}, and (3) the PRISM rows in the system-level comparison in Table~\ref{tab:prior_comparison_full}.

\subsection{Family-level configurations}
\label{app:family_configs}

Table~\ref{tab:family_configs} reports the configurations used for the family-level leaderboard. CaptainCook4D and EgoPER configurations were selected by validation Any-F1; Type-F1, AUC, and Seq break ties in that order. EgoOops uses the configuration with the highest mean validation Any-F1 across CaptainCook4D and EgoPER for each representation family. EgoOops error-video labels do not enter selection. All test metrics in a leaderboard row come from the same configuration.

\begin{table*}[t]
\centering
\small
\setlength{\tabcolsep}{4pt}
\begin{tabular}{llccccccc}
\toprule
\textbf{Dataset} & \textbf{Family} & \textbf{Encoder} & \textbf{$d$} & \textbf{Norm} & \textbf{Emission} & \textbf{Temp} & \textbf{Align} \\
\midrule

\multirow{7}{*}{EgoOops}
& Vision           & V-JEPA           & 1024 & L2   & Gaussian & -- & NW \\
& Gold text        & CLIP             & 128  & none & cosine   & 6  & NW \\
& Gold + GPT-SRL           & CLIP             & 128  & none & cosine   & 12 & difflib \\
& Qwen text             & Instructor-XL    & 64   & L2   & Gaussian & -- & difflib \\
& Qwen + GPT-SRL    & Instructor-XL    & 64   & none & Gaussian & -- & difflib \\
& Gemini text             & all-MPNet        & 768  & L2   & Gaussian & -- & difflib \\
& Gemini + GPT-SRL  & Instructor-XL    & 64   & none & Gaussian & -- & difflib \\
\midrule

\multirow{7}{*}{EgoPER}
& Vision           & VideoMAE-v2      & 128  & L2   & cosine   & 6  & difflib \\
& Gold text        & CLIP             & 256  & none & Gaussian & -- & difflib \\
& Gold + GPT-SRL & RoBERTa-SimCSE   & 256  & L2   & Gaussian & -- & NW \\
& Qwen text             & CLIP             & 256  & L2   & Gaussian & -- & NW \\
& Qwen + GPT-SRL    & RoBERTa-SimCSE   & 256  & none & Gaussian & -- & NW \\
& Gemini text             & CLIP             & 256  & L2   & Gaussian & -- & NW \\
& Gemini + GPT-SRL  & RoBERTa-SimCSE   & 128  & L2   & Gaussian & -- & NW \\
\midrule

\multirow{7}{*}{CaptainCook4D}
& Vision           & VideoMAE-v2      & 768  & none & cosine   & 6  & difflib \\
& Gold text        & RoBERTa-SimCSE   & 256  & none & cosine   & 10 & NW \\
& Gold + GPT-SRL  & RoBERTa-SimCSE   & 256  & none & cosine   & 10 & difflib \\
& Qwen text             & all-MiniLM       & 256  & L2   & cosine   & 12 & difflib \\
& Qwen + GPT-SRL    & RoBERTa-SimCSE   & 256  & L2   & cosine   & 10 & difflib \\
& Gemini text             & RoBERTa-SimCSE   & 128  & none & cosine   & 12 & difflib \\
& Gemini + GPT-SRL  & RoBERTa-SimCSE   & 256  & none & cosine   & 10 & difflib \\
\bottomrule
\end{tabular}
\caption{Configurations used for the family-level leaderboard in Table~\ref{tab:main_results}. CaptainCook4D and EgoPER configurations are selected by validation Any-F1. EgoOops configurations are transferred from the best mean validation result across these two datasets. EgoOops error-video labels do not enter selection.}
\label{tab:family_configs}
\end{table*}

\subsection{Representation comparison configurations}
\label{app:repr_grounding_configs}

Table~\ref{tab:repr_grounding_configs} lists the fixed downstream
configurations for the matched Raw--GPT-SRL comparisons in
Table~\ref{tab:repr_grounding}. Each configuration applies to both VLMs
within one dataset.

\begin{table*}[t]
\centering
\small
\setlength{\tabcolsep}{4pt}
\begin{tabular}{llcccccl}
\toprule
\bf Dataset & \bf Encoder & \bf Output $d$ & \bf PCA $d$ &
\bf Norm & \bf Emission & \bf Temp. & \bf Align \\
\midrule
EgoOops & Instructor-XL
& 768 & 128 & L2 & cosine & 8 & NW \\
EgoPER & RoBERTa-SimCSE
& 1024 & 128 & L2 & Gaussian & -- & NW \\
CaptainCook4D & RoBERTa-SimCSE
& 1024 & 256 & L2 & cosine & 10 & difflib \\
\bottomrule
\end{tabular}
\caption{Fixed downstream configurations for the matched Raw--GPT-SRL
comparisons and decoder controls. Output $d$ is the encoder output
dimension; PCA $d$ is the dimension after projection.}
\label{tab:repr_grounding_configs}
\end{table*}

\subsection{Prior-work comparison configurations}
\label{app:prior_configs}

Table~\ref{tab:prior_configs} reports the configurations used for the PRISM rows in Table~\ref{tab:prior_comparison_full}. CaptainCook4D and EgoPER configurations were selected by validation EDA. EgoPER's validation split contains only correct videos, so its validation EDA measures the proportion of correct segments accepted as normal and does not measure error recall. Test metrics were not used for selection. All benchmark-specific metrics within each row come from the same configuration.

\begin{table*}[t]
\centering
\small
\setlength{\tabcolsep}{4pt}
\begin{tabular}{lllccccl}
\toprule
\bf Dataset & \bf VLM & \bf Representation & \bf Encoder &
\bf PCA $d$ & \bf Norm & \bf Emission & \bf Align \\
\midrule
\multirow{4}{*}{EgoPER}
& Qwen   & Raw     & all-MiniLM     & 256 & L2   & Gaussian & difflib \\
& Qwen   & GPT-SRL & RoBERTa-SimCSE & 64  & none & Gaussian & difflib \\
& Gemini & Raw     & all-MiniLM     & 256 & L2   & Gaussian & difflib \\
& Gemini & GPT-SRL & RoBERTa-SimCSE & 64  & none & Gaussian & NW \\
\midrule
\multirow{4}{*}{CaptainCook4D}
& Qwen   & Raw     & RoBERTa-SimCSE & 256 & none & Gaussian & difflib \\
& Qwen   & GPT-SRL & RoBERTa-SimCSE & 256 & L2   & Gaussian & difflib \\
& Gemini & Raw     & RoBERTa-SimCSE & 256 & L2   & Gaussian & difflib \\
& Gemini & GPT-SRL & RoBERTa-SimCSE & 256 & L2   & Gaussian & difflib \\
\bottomrule
\end{tabular}
\caption{Configurations used for the PRISM rows in the system-level comparison in Table~\ref{tab:prior_comparison_full}. PCA $d$ is the projected dimensionality.}
\label{tab:prior_configs}
\end{table*}

\section{Additional results and diagnostics}
\label{app:additional_ablations}

\subsection{Grounding accuracy}
\label{app:grounding_exact}

Table~\ref{tab:grounding_exact} reports Exact, the proportion of evaluated segments whose generated description matches the target description word for word. Seq instead measures Step-ID recovery after Needleman--Wunsch alignment, as defined in Section~\ref{sec:exp}. A non-verbatim description can therefore fail Exact while still producing the correct Step ID. Table~\ref{tab:stratified_srl} reports positionwise Step-ID accuracy within grounding strata, without sequence alignment.

\begin{table*}[t]
\begin{center}
\small
\setlength{\tabcolsep}{5pt}
\begin{tabular}{lccc}
\toprule
\bf Dataset & \bf Qwen Exact & \bf Gemini Exact & \bf $\Delta$ \\
\midrule
EgoOops        & 0.425 & \textbf{0.496} & +0.070 \\
EgoPER         & 0.806 & \textbf{0.864} & +0.057 \\
CaptainCook4D  & 0.578 & \textbf{0.695} & +0.117 \\
\bottomrule
\end{tabular}
\end{center}
\caption{Word-for-word grounding accuracy before Step-ID decoding. $\Delta$ is Gemini minus Qwen; differences are computed before rounding.}
\label{tab:grounding_exact}
\end{table*}

\subsection{Second SRL generator}
\label{app:qwen_srl_text}

Table~\ref{tab:qwen_srl_matched} repeats the matched grounded comparison with Qwen3.6-27B SRLs. Qwen-SRL improves five of six Raw conditions. GPT-SRL gives higher Any-F1 in five conditions. The two generators tie after rounding on CaptainCook4D with Gemini grounding. Gains occur with both SRL generators, although their magnitude varies across conditions.

\begin{table*}[t]
\centering
\small
\setlength{\tabcolsep}{7pt}
\begin{tabular}{llccc}
\toprule
\bf Dataset & \bf Grounding & \bf Raw & \bf GPT-SRL & \bf Qwen-SRL \\
\midrule
\multirow{2}{*}{EgoOops}
& Qwen   & 0.321 & \textbf{0.356} & 0.341 \\
& Gemini & 0.349 & \textbf{0.353} & 0.341 \\
\midrule
\multirow{2}{*}{EgoPER}
& Qwen   & 0.578 & \textbf{0.601} & 0.584 \\
& Gemini & 0.602 & \textbf{0.631} & 0.613 \\
\midrule
\multirow{2}{*}{CaptainCook4D}
& Qwen   & 0.587 & \textbf{0.592} & 0.588 \\
& Gemini & 0.595 & \textbf{0.596} & \textbf{0.596} \\
\bottomrule
\end{tabular}
\caption{Matched Any-F1 with GPT-5.3 and Qwen3.6-27B SRLs.
Every row uses the fixed downstream configuration reported in
Table~\ref{tab:repr_grounding_configs}.}
\label{tab:qwen_srl_matched}
\end{table*}

\subsection{Stratified SRL analysis}
\label{app:stratified_srl}

Table~\ref{tab:stratified_srl} stratifies the matched Viterbi outputs to locate the Step-ID changes caused by GPT-SRL. For this diagnostic, each decoded ID is compared with the annotated ID of the same input segment before sequence alignment. Exact contains word-for-word target matches. Same ID contains non-verbatim descriptions associated with the annotated ID. Other ID contains descriptions associated with another ontology ID. Ambiguous/open contains strings associated with multiple IDs or absent from the ontology. Modified is an overlapping subset whose target annotation contains a modified description. The analysis uses annotations only to define diagnostic strata. This positionwise diagnostic differs from the alignment-aware Seq metric reported in Table~\ref{tab:greedy_decoder_ablation}.

\begin{table*}[t]
\centering
\small
\renewcommand{\arraystretch}{1.12}
\begin{tabular*}{\textwidth}{@{\extracolsep{\fill}}llccccc@{}}
\toprule
\bf Dataset & \bf VLM & \bf Exact & \bf Same ID & \bf Other ID
& \bf Ambiguous/open & \bf Modified \\
\midrule
EgoOops
& Qwen   & +0.021 (145) & +0.095 (21) & -0.006 (170)
& -0.200 (5) & +0.034 (59) \\
EgoOops
& Gemini & +0.000 (169) & +0.000 (19) & +0.020 (153)
& -- & +0.000 (59) \\
\midrule
EgoPER
& Qwen   & +0.014 (1771) & -0.008 (121) & +0.000 (284)
& +0.250 (20) & +0.054 (500) \\
EgoPER
& Gemini & +0.006 (1897) & +0.000 (104) & -0.005 (182)
& +0.077 (13) & +0.024 (500) \\
\midrule
CaptainCook4D
& Qwen   & -0.004 (891) & -0.013 (80) & -0.013 (550)
& +0.048 (21) & -0.006 (363) \\
CaptainCook4D
& Gemini & +0.014 (1071) & +0.000 (130) & +0.013 (318)
& -0.087 (23) & +0.025 (363) \\
\bottomrule
\end{tabular*}
\caption{Change in positionwise decoded-ID accuracy from Raw to GPT-SRL within each grounding stratum, before sequence alignment. Support appears in parentheses. Modified overlaps the first four strata. A dash marks an empty stratum.}
\label{tab:stratified_srl}
\end{table*}

\subsection{Performance by error type}
\label{app:per_error}

Table~\ref{tab:per_error} reports type-specific F1 and support for all twelve matched grounded conditions in Table~\ref{tab:repr_grounding}. The Support columns report $TP+FN$. A single value applies to both Raw and GPT-SRL. Raw/GPT-SRL values are shown when support differs after reference selection. EgoOops has no gold Deletion or Transposition instances. Their F1 is zero under the evaluator's zero-division convention.

For each decoded Step-ID sequence, the evaluator selects the valid reference with the lowest alignment cost; the number of ordering inversions breaks ties. The selected reference defines both the predicted structural edits and the structural targets for that condition. Different predictions may therefore select different valid references, so structural-error support may vary across conditions. Insertions and deletions correspond to unmatched executed and reference steps. A transposition pairs an insertion and deletion of the same Step ID at different alignment positions. Substitution targets come from the benchmark annotations after the mapping in Appendix~\ref{app:label_mapping}. Any-F1 uses the same reference-selection and error-construction procedure.

\begin{table*}[t]
\centering
\small
\setlength{\tabcolsep}{5pt}
\begin{tabular}{llccc|ccc}
\toprule
& & \multicolumn{3}{c|}{\bf Qwen grounding}
  & \multicolumn{3}{c}{\bf Gemini grounding} \\
\cmidrule(lr){3-5}\cmidrule(lr){6-8}
\bf Dataset & \bf Error type
& \bf Raw & \bf GPT-SRL & \bf Support
& \bf Raw & \bf GPT-SRL & \bf Support \\
\midrule

\multirow{4}{*}{EgoOops}
& Insertion     & 0.082 & 0.119 & 89 & 0.151 & 0.170 & 90 \\
& Deletion      & 0.000 & 0.000 & 0  & 0.000 & 0.000 & 0  \\
& Substitution  & 0.220 & 0.215 & 62 & 0.164 & 0.177 & 62 \\
& Transposition & 0.000 & 0.000 & 0  & 0.000 & 0.000 & 0  \\

\midrule
\multirow{4}{*}{EgoPER}
& Insertion     & 0.203 & 0.189 & 199/195 & 0.185 & 0.208 & 209/207 \\
& Deletion      & 0.742 & 0.720 & 307     & 0.778 & 0.777 & 306     \\
& Substitution  & 0.349 & 0.430 & 354     & 0.384 & 0.442 & 354     \\
& Transposition & 0.093 & 0.117 & 66/64   & 0.100 & 0.133 & 53/57   \\

\midrule
\multirow{4}{*}{CaptainCook4D}
& Insertion     & 0.000 & 0.000 & 4       & 0.000 & 0.000 & 4       \\
& Deletion      & 0.167 & 0.218 & 71      & 0.208 & 0.257 & 71      \\
& Substitution  & 0.137 & 0.146 & 238     & 0.146 & 0.175 & 238     \\
& Transposition & 0.279 & 0.281 & 366     & 0.298 & 0.306 & 366/367 \\
\bottomrule
\end{tabular}
\caption{Type-specific F1 and evaluation support for all matched grounded conditions in Table~\ref{tab:repr_grounding}. Support is $TP+FN$. Raw/GPT-SRL values are shown when support differs between conditions.}
\label{tab:per_error}
\end{table*}

\subsection{Bootstrap confidence intervals}
\label{app:bootstrap}

We estimate uncertainty for the six matched Raw--SRL comparisons in Table~\ref{tab:repr_grounding} with a paired video-level bootstrap. We sample test videos with replacement and recompute Any-F1 from the pooled step-level counts. Raw and GPT-SRL use the same sampled videos in each of 10,000 replicates. Table~\ref{tab:bootstrap} reports percentile 95\% confidence intervals for both conditions and their paired difference.

The interval for the SRL gain excludes zero for Qwen on EgoOops and for both VLMs on EgoPER. It includes zero for Gemini on EgoOops and for both CaptainCook4D comparisons. The point estimate remains positive in all six comparisons. These intervals quantify uncertainty within PRISM. They do not establish statistical superiority over the published systems in Table~\ref{tab:prior_comparison_full}, whose per-video predictions are unavailable.

\begin{table*}[t]
\centering
\small
\setlength{\tabcolsep}{6pt}
\renewcommand{\arraystretch}{1.08}
\begin{tabular}{llccc}
\toprule
\bf Dataset & \bf VLM
& \bf Raw Any-F1 [95\% CI]
& \bf GPT-SRL Any-F1 [95\% CI]
& $\boldsymbol{\Delta}$ \bf [95\% CI] \\
\midrule
EgoOops
& Qwen
& 0.321 [0.252, 0.380]
& 0.356 [0.295, 0.407]
& +0.036 [+0.005, +0.076] \\
& Gemini
& 0.349 [0.287, 0.414]
& 0.353 [0.279, 0.426]
& +0.003 [$-$0.028, +0.035] \\
\midrule
EgoPER
& Qwen
& 0.578 [0.544, 0.611]
& 0.601 [0.565, 0.638]
& +0.024 [+0.007, +0.041] \\
& Gemini
& 0.602 [0.569, 0.633]
& 0.631 [0.594, 0.668]
& +0.030 [+0.011, +0.048] \\
\midrule
CaptainCook4D
& Qwen
& 0.587 [0.520, 0.643]
& 0.592 [0.525, 0.650]
& +0.005 [$-$0.007, +0.017] \\
& Gemini
& 0.595 [0.530, 0.649]
& 0.596 [0.533, 0.649]
& +0.001 [$-$0.013, +0.015] \\
\bottomrule
\end{tabular}
\caption{Paired video-level bootstrap 95\% confidence intervals for Any-F1 in the six matched grounded comparisons. We use 10,000 replicates and percentile intervals. $\Delta$ is GPT-SRL minus Raw; each replicate uses identical sampled video IDs for both conditions. Differences are computed before rounding.}
\label{tab:bootstrap}
\end{table*}

\subsection{Multimodal fusion sweep}
\label{app:mm_sweep}

Table~\ref{tab:mm_sweep} reports the multimodal weight sweep. In the multimodal setting, we combine a visual step embedding with its paired text embedding by weighted interpolation, sweeping from text-dominant to vision-dominant mixtures over a fixed grid of weights to test how much visual signal can be added before HMM sequence recovery and mistake recognition degrade.

The result pattern is consistent with the family-level results. Text-heavy mixtures are strongest on EgoOops. On CaptainCook4D, increasing the visual weight gives a small Any-F1 gain but reduces Seq and AUC. On EgoPER, larger visual weight raises Seq and AUC but leaves Any-F1 nearly flat. We therefore do not present multimodal fusion as the universal solution. In this HMM setting, textual embeddings anchor the state space more sharply than frozen visual features.

\begin{table*}[t]
\begin{center}
\small
\setlength{\tabcolsep}{2pt}
\begin{tabularx}{\textwidth}{@{}>{\raggedright\arraybackslash}p{0.20\textwidth}*{9}{C}@{}}
\toprule
& \multicolumn{3}{c}{\bf EgoOops} & \multicolumn{3}{c}{\bf EgoPER} & \multicolumn{3}{c}{\bf CC4D} \\
\cmidrule(lr){2-4}\cmidrule(lr){5-7}\cmidrule(lr){8-10}
\bf Weights $(w_{\mathrm{img}}, w_{\mathrm{txt}})$ & \bf Seq & \bf Any-F1 & \bf AUC & \bf Seq & \bf Any-F1 & \bf AUC & \bf Seq & \bf Any-F1 & \bf AUC \\
\midrule
$(0.2, 0.8)$ & 0.651 & 0.437 & 0.558 & 0.235 & 0.478 & 0.441 & 0.598 & 0.547 & 0.604 \\
$(0.4, 0.6)$ & 0.651 & 0.429 & 0.562 & 0.479 & 0.474 & 0.558 & 0.569 & 0.550 & 0.576 \\
$(0.6, 0.4)$ & 0.651 & 0.429 & 0.563 & 0.629 & 0.463 & 0.597 & 0.538 & 0.554 & 0.561 \\
$(0.8, 0.2)$ & 0.651 & 0.429 & 0.562 & 0.700 & 0.466 & 0.599 & 0.502 & 0.558 & 0.549 \\
\bottomrule
\end{tabularx}
\end{center}
\caption{Multimodal fusion sweep over interpolation weights between visual and textual step embeddings. For each dataset, we fix one visual--textual configuration and vary only the fusion weights.}
\label{tab:mm_sweep}
\end{table*}

\subsection{Multimodal results with Qwen3.6-27B SRL}
\label{app:qwen_srl_mm}

Table~\ref{tab:qwen_srl_mm_matched} compares multimodal runs that use GPT-SRL and Qwen-SRL. 
The runs are matched by dataset, VLM text source, visual encoder, feature dimension, fusion weights, emission model, and alignment method.
The table reports mean deltas, computed as Qwen-SRL minus GPT-SRL.

Across the matched multimodal runs, the mean differences between GPT-SRL and Qwen-SRL are close to zero on all three datasets. Neither generator has a consistent average advantage under the evaluated configurations. Individual configurations may respond differently.

\begin{table*}[t]
\begin{center}
\small
\setlength{\tabcolsep}{5pt}
\begin{tabular}{lccccc}
\toprule
\bf Dataset & \bf Matched runs & \bf $\Delta$ Seq & \bf $\Delta$ Any-F1 & \bf $\Delta$ Type-F1 & \bf $\Delta$ AUC \\
\midrule
EgoOops        & 160 & -0.0003 & -0.0003 & +0.0016 & -0.0002 \\
EgoPER         & 240 & +0.0007 & -0.0014 & -0.0005 & +0.0005 \\
CaptainCook4D  & 240 & +0.0007 & +0.0002 & -0.0006 & -0.0004 \\
\bottomrule
\end{tabular}
\end{center}
\caption{Matched multimodal comparison of GPT-SRL and Qwen-SRL. Deltas are computed as Qwen-SRL minus GPT-SRL. AUC denotes Step-AUC.}
\label{tab:qwen_srl_mm_matched}
\end{table*}

\section{Additional qualitative examples}
\label{app:qual_examples}

This appendix illustrates the qualitative patterns behind the main quantitative results. We focus on three questions: when SRL helps over grounded text without SRL, what the action--object abstraction preserves and loses, and how the unified edit-based view applies across datasets.

\subsection{Error-relevant semantic distinctions}

Table~\ref{tab:srl_help_examples} lists error-relevant distinctions that SRL can represent explicitly, including relations, destinations, materials, and nested events.

\begin{table*}[t]
\begin{center}
\small
\setlength{\tabcolsep}{2pt}
\renewcommand{\arraystretch}{1.1}
\begin{tabularx}{\textwidth}{@{}>{\raggedright\arraybackslash}p{0.14\textwidth}YYY@{}}
\toprule
\textbf{Dataset} & \textbf{Reference step} & \textbf{Observed / grounded variant} & \textbf{Semantic distinction} \\
\midrule
\multirow{4}{*}{CC4D}
& Place the egg from the cup over the lettuce
& Place the egg from the cup under the lettuce
& Makes the relation contrast explicit: \texttt{over} vs.\ \texttt{under}. \\

& Invert the mug to release the cake onto a plate
& Invert the mug to release the cake onto a bowl \newline
Invert the mug to release the cake onto an unclean plate
& Isolates destination and destination-property changes while preserving the same event structure. \\

& Invert the mug to release the cake onto a plate
& Use a spoon to invert the mug to release the cake onto a plate
& Preserves the added instrumental subevent as nested structure. \\

& Measure and add 1.5 tbsp sugar to the mixing bowl
& Qwen: Measure and add 1.5 tbsp brown sugar to the mixing bowl
& Preserves the \texttt{MEASURE \& ADD} structure and highlights the ingredient mismatch. \\

\midrule
\multirow{1}{*}{EgoOops}
& Put three zinc plates on the center column of the microplate using the pair of tweezers
& Qwen: Put three copper plates on the left column of the microplate using the pair of tweezers
& Makes both material and location drift explicit. \\
\bottomrule
\end{tabularx}
\end{center}
\caption{Examples of error-relevant distinctions that SRL can represent explicitly, including relations, destinations, properties, materials, and nested events. CC4D denotes CaptainCook4D.}
\label{tab:srl_help_examples}
\end{table*}

\subsection{What action--object preserves and loses}

Table~\ref{tab:actobj_examples} shows what the action--object abstraction preserves and what it weakens. Our action--object representation retains the main action together with all visually involved objects, not only the semantic patient. Tools and contacted objects are therefore preserved, which makes the representation useful for some mistakes but too coarse for others.

\begin{table*}[t]
\begin{center}
\small
\setlength{\tabcolsep}{2pt}
\renewcommand{\arraystretch}{1.1}
\begin{tabularx}{\textwidth}{@{}>{\raggedright\arraybackslash}p{0.12\textwidth}
                              >{\raggedright\arraybackslash}p{0.12\textwidth}
                              YYY@{}}
\toprule
\textbf{Effect} & \textbf{Dataset} & \textbf{Reference step} & \textbf{Contrasting step} & \textbf{What action--object captures or loses} \\
\midrule
\multirow{2}{*}{Preserves}
& EgoPER
& Use knife to scoop jelly
& Use spoon to scoop jelly
& Preserves the tool distinction because tools remain in the object set. \\

& CC4D
& Sprinkle 1 tablespoon of cheese on cup
& Pour 1 tablespoon of cheese on cup
& Preserves the predicate change. \\

\midrule
\multirow{4}{*}{Weakens}
& CC4D
& Microwave just until cheese melts, about 10 seconds
& Microwave for 30 seconds
& Loses duration and completion-condition information. \\

& CC4D
& Top cup with 1 tablespoon of salsa
& Top cup with 1/2 tablespoon of salsa
& Loses scalar quantity information. \\

& EgoPER
& Slowly pour the rest of water in circular motion
& --
& Loses manner information. \\

& CC4D
& Replace the top of the English muffin
& Flip over the top muffin and replace it on top
& Weakens multi-action structure. \\
\bottomrule
\end{tabularx}
\end{center}
\caption{Qualitative examples of what the action--object abstraction preserves and what it weakens. It remains informative for predicate and tool changes, but loses detail when the mistake depends on duration, scalar quantity, manner, or multi-action structure. CC4D denotes CaptainCook4D.}
\label{tab:actobj_examples}
\end{table*}

EgoPER contributes a different qualitative pattern from EgoOops and CaptainCook4D. For EgoPER, the VLM receives the task-specific candidate-step vocabulary as grounding context. Its short and regular candidate descriptions limit lexical variation, so Raw and SRL remain close on this dataset. The more informative contrast in EgoPER may appear between richer textual representations and the compressed action--object abstraction, especially for manner-sensitive steps such as \texttt{Slowly pour the rest of water in circular motion}, \texttt{POUR(Agent: you, Object: rest(of(water)), Manner: slowly \& in(circular\_motion))}.

\subsection{Examples of the unified error taxonomy}

Table~\ref{tab:taxonomy_examples} shows how the unified taxonomy matches the procedural structure of the observed deviations across datasets.

\begin{table*}[t]
\begin{center}
\small
\setlength{\tabcolsep}{2pt}
\renewcommand{\arraystretch}{1.1}
\begin{tabularx}{\textwidth}{@{}>{\raggedright\arraybackslash}p{0.14\textwidth}
                              >{\raggedright\arraybackslash}p{0.10\textwidth}
                              YY@{}}
\toprule
\textbf{Type} & \textbf{Dataset} & \textbf{Example} & \textbf{Why it belongs to this type} \\
\midrule
\multirow{3}{*}{Insertion}
& EgoOops
& Extra grasp actions such as \texttt{Grasp black light} or \texttt{Grasp a long blue thin cuboid block}
& Extra steps outside the successful reference sequence. \\

& EgoPER
& \texttt{Pour water from the second bowl to the first one}
& Corrective action inserted after an earlier mistake. \\

& EgoOops
& \texttt{Mix all the liquid and shine a black light on it}
& Composite action inserted outside the reference sequence. \\

\midrule
Deletion
& EgoPER
& \texttt{Put bananas} annotated as \texttt{Skipped this step}
& Reference step missing from the observed sequence. \\

\midrule
\multirow{3}{*}{Substitution}
& CC4D
& \texttt{Place the egg from the cup over the lettuce} vs.\ \texttt{... under the lettuce}
& Same step slot, wrong relation. \\

& CC4D
& \texttt{Invert the mug to release the cake onto a plate} vs.\ \texttt{... onto a bowl}
& Same step slot, wrong destination. \\

& CC4D
& \texttt{Measure and add 1.5 tbsp sugar to the mixing bowl} vs.\ \texttt{... 1.5 tbsp brown sugar ...}
& Same step structure, wrong ingredient. \\

\midrule
Transposition
& CC4D
& \texttt{Sprinkle 2 tablespoon of cheese on cup} is performed before \texttt{Microwave for 10 seconds}
& Wrong local order of otherwise valid steps. \\
\bottomrule
\end{tabularx}
\end{center}
\caption{Qualitative examples of the unified edit-based taxonomy across benchmarks. The same four types remain interpretable despite different dataset-specific error labels and annotation styles. CC4D denotes CaptainCook4D.}
\label{tab:taxonomy_examples}
\end{table*}

\section{Decoder Ablations}
\label{app:ablations}

We use the matched inputs and alignment methods from Table~\ref{tab:repr_grounding_configs}. The nearest--greedy comparison adds PCA and the trained emission model. The greedy--Viterbi comparison changes only the decoding rule.

\paragraph{Independent nearest-candidate matching (no HMM).}
Let $x_t$ denote the description generated by the VLM for video segment $t$. Let $\mathcal{S}_a$ denote the candidate step set for task $a$, and let $d_s$ denote the canonical description of step $s$. In the GPT-SRL condition, both $x_t$ and $d_s$ are replaced by their semantic role representations.

The text encoder $e$ encodes the generated description and every candidate description. All embeddings are L2-normalized. Each segment is assigned independently to the candidate with the highest cosine similarity:
\[
\hat{s}_t
=
\arg\max_{s \in \mathcal{S}_a}
\cos\!\left(e(x_t), e(d_s)\right).
\]

This decoder uses no HMM checkpoint, start probabilities, transition probabilities, or trained emission parameters. It assigns each Step ID without information from adjacent segments. Candidate cosine similarity supplies the LowProbEmission score; LowProbTransition is disabled because this decoder has no transition model. The semantic-substitution check compares the observation with the assigned candidate embedding rather than a learned state mean. The predicted sequence $\hat{s}_{1:T}$ enters the same edit-alignment and error-detection modules as PRISM. Sequence information enters only at the alignment stage.

This baseline isolates the performance provided by the frozen language representation, the task-specific candidate vocabulary, and edit alignment, without any trained HMM component. The Raw--SRL difference measures the contribution of SRL without HMM training or sequence priors. The procedure is deterministic once the VLM descriptions and text encoder are fixed.

\paragraph{Greedy emission decoding (without transition modeling).}
PRISM obtains each observation vector as
\[
z_t=P\!\left(N(e(x_t))\right),
\]
where $e$ is the text encoder, $N$ is the configured normalization, and $P$ is the fitted PCA projection. It then computes an emission score $E_{t,s}$ for every segment $t$ and state $s$. The complete HMM applies the Viterbi objective in Equation~\ref{eq:23} to the emission matrix $\bm{E}$.

The greedy decoder uses the same observation vectors and the same trained emission model. It assigns each segment independently:
\[
\hat{s}_t^{\mathrm{greedy}}
=
\arg\max_{s \in \mathcal{S}_a} E_{t,s}.
\]
In implementation, this operation applies a row-wise argmax to the emission matrix $E\in\mathbb{R}^{T\times|\mathcal{S}_a|}$. The decoder removes the start probabilities, transition probabilities, and Viterbi search. It retains the encoder, PCA projection, normalization, trained emission parameters, and all downstream settings. The resulting Step ID sequence is passed to the same edit-alignment module.

This ablation isolates the contribution of global sequence decoding. The comparison between nearest-candidate matching and greedy decoding measures the combined contribution of the fitted PCA projection and trained local emission model. The comparison between greedy and Viterbi decoding measures the contribution of the sequence prior.

\paragraph{Configurations and evaluation.}
The nearest-candidate decoder uses the matched text encoder and alignment method. It operates on the native encoder embeddings with L2 normalization and raw cosine similarity. PCA dimension, HMM emission family, and emission temperature do not apply to this decoder.

The greedy decoder uses the complete matched PRISM configurations reported in Table~\ref{tab:repr_grounding_configs}. All settings remain fixed except the decoding rule.

Greedy decoding improves Any-F1 over nearest-candidate matching in all twelve conditions. Viterbi improves Any-F1 for both VLMs on CaptainCook4D and in three EgoOops conditions. It leaves all four EgoPER conditions unchanged: Gaussian emissions already select the same Step IDs before sequence decoding. It lowers Qwen Raw on EgoOops from 0.339 to 0.321, where transition estimation uses 21 correct videos across five tasks. On CaptainCook4D, Viterbi improves Any-F1 and lowers Seq. Seq measures Step-ID recovery; Any-F1 measures mistake
recognition after alignment.

\makeatletter
\setlength{\@fptop}{0pt}
\setlength{\@dblfptop}{0pt}
\makeatother

\begin{table*}[p]
\centering
\small
\setlength{\tabcolsep}{6pt}
\renewcommand{\arraystretch}{1.15}

\begin{tabular*}{\textwidth}{
@{\extracolsep{\fill}}
lllcc|cc|cc
@{}
}
% \begin{table*}[t]
% \centering
% \small
% \renewcommand{\arraystretch}{1.1}
% \begin{tabular*}{\textwidth}{@{\extracolsep{\fill}}lllcc|cc|cc@{}}
\toprule
& & & \multicolumn{2}{c|}{\bf Nearest} & \multicolumn{2}{c|}{\bf Greedy} & \multicolumn{2}{c}{\bf Viterbi} \\
\cmidrule(lr){4-5}\cmidrule(lr){6-7}\cmidrule(lr){8-9}
\bf Dataset & \bf VLM & \bf Repr. & \bf Seq & \bf Any-F1 & \bf Seq & \bf Any-F1 & \bf Seq & \bf Any-F1 \\
\midrule
\multirow{4}{*}{EgoOops}
& Qwen   & Raw     & 0.501 & 0.337 & 0.501 & 0.339 & 0.575 & 0.321 \\
& Qwen   & GPT-SRL & 0.499 & 0.335 & 0.501 & 0.351 & 0.587 & 0.356 \\
& Gemini & Raw     & 0.560 & 0.305 & 0.572 & 0.339 & 0.619 & 0.349 \\
& Gemini & GPT-SRL & 0.560 & 0.315 & 0.569 & 0.328 & 0.613 & 0.353 \\
\midrule
\multirow{4}{*}{EgoPER}
& Qwen   & Raw     & 0.803 & 0.547 & 0.803 & 0.578 & 0.803 & 0.578 \\
& Qwen   & GPT-SRL & 0.816 & 0.524 & 0.816 & 0.601 & 0.816 & 0.601 \\
& Gemini & Raw     & 0.849 & 0.571 & 0.849 & 0.602 & 0.849 & 0.602 \\
& Gemini & GPT-SRL & 0.854 & 0.557 & 0.854 & 0.631 & 0.854 & 0.631 \\
\midrule
\multirow{4}{*}{CaptainCook4D}
& Qwen   & Raw     & 0.630 & 0.567 & 0.631 & 0.581 & 0.499 & 0.587 \\
& Qwen   & GPT-SRL & 0.630 & 0.559 & 0.630 & 0.578 & 0.485 & 0.592 \\
& Gemini & Raw     & 0.782 & 0.538 & 0.783 & 0.564 & 0.542 & 0.595 \\
& Gemini & GPT-SRL & 0.785 & 0.528 & 0.783 & 0.561 & 0.555 & 0.596 \\
\bottomrule
\end{tabular*}
\caption{Decoder ablations under the matched configurations. Nearest-candidate matching uses no HMM. Greedy decoding retains the trained local decoder. Viterbi adds start and transition probabilities and global sequence search.}
\label{tab:greedy_decoder_ablation}
\end{table*}

\end{document}